\relax
\documentclass[letterpaper]{article} 
\usepackage{aaai22}  
\usepackage{times}  
\usepackage{helvet}  
\usepackage{courier}  
\usepackage[hyphens]{url}  
\usepackage{graphicx} 
\usepackage{natbib}  
\usepackage{caption} 
\DeclareCaptionStyle{ruled}{labelfont=normalfont,labelsep=colon,strut=off} 
\usepackage{amsmath,amssymb,amsfonts}
\usepackage{amsthm}
\usepackage{bbm}
\usepackage{algorithm}
\usepackage[noend]{algpseudocode}
\usepackage{subfigure}
\usepackage{multirow}

\usepackage{hyperref} 
\usepackage{listings}

\newtheorem{theorem}{Theorem}
\newtheorem{definition}{Definition}

\newtheorem{lemma}{Lemma}
\newtheorem{corollary}{Corollary}

\newtheorem{assumption}{Assumption}
\newcommand{\system}{LGC}

\newcommand{\ie}{\textit{i}.\textit{e}.}
\newcommand{\eg}{\textit{e}.\textit{g}.}

\newcommand{\para}[1]{\noindent {\bf #1}}

\usepackage{newfloat}
\usepackage{listings}
\usepackage{bibentry}
\floatstyle{ruled}
\newfloat{listing}{tb}{lst}{}
\floatname{listing}{Listing}

\title{Toward Efficient Federated Learning in Multi-Channeled Mobile Edge Network with Layerd Gradient Compression}

 \author{ 
     Haizhou Du\textsuperscript{\rm 1},
 Xiaojie Feng\textsuperscript{\rm 1},
   Qiao Xiang\textsuperscript{\rm 2}\equalcontrib,
     Haoyu Liu\textsuperscript{\rm 2}
 }
 \affiliations{
     \textsuperscript{\rm 1}Shanghai University of Electric Power,

       \textsuperscript{\rm 2}Xiamen University

}

\begin{document}

\maketitle

\begin{abstract}
    A fundamental issue for federated learning (FL) is how to achieve optimal model performance under highly dynamic communication environments. This issue can be alleviated by the fact that modern edge devices usually can connect to the edge FL server via multiple communication channels (\eg, 4G, LTE and 5G).  However, having an edge device send copies of local models to the FL server along multiple channels is redundant, time-consuming, and would waste resources (\eg, bandwidth, battery life and monetary cost). In this paper, motivated by the layered coding techniques in video streaming, we propose a novel FL framework called layered gradient  compression (LGC). Specifically, in LGC, local gradients from a device is coded into several layers and each layer is sent to the FL server along a different channel. The FL server aggregates the received layers of local gradients from devices to update the global model, and sends the result back to the devices.  We prove the convergence of LGC, and formally define the problem of resource-efficient federated learning with LGC. We then propose a learning-based algorithm for each device to dynamically adjust its local computation (\ie, the number of local stochastic descent) and communication decisions (\ie, the compression level of different layers and the layer-to-channel mapping) in each iteration. Results from extensive experiments show that using our algorithm, LGC significantly reduces the training time, improves the resource utilization, while achieving a similar accuracy, compared with well-known FL mechanisms.
\end{abstract}
\section{Introduction}

Federated learning (FL) has emerged as an efficient solution to analyze and process distributed data for data-driven tasks (\eg, autonomous driving, virtual reality, image classification, etc.) in Mobile Edge Computing (MEC) \cite{niknam2020federated, li2021privacy,verma2018federated,wang2018edge,yang2019federated}. By performing training tasks at edge devices (\eg, mobile phones and tablets) and aggregating the learned parameters at edge servers, FL significantly reduces the network bandwidth usage of machine learning applications, and protects the data privacy of edge devices~\cite{bonawitz2019towards}. 


However, to practically deploy FL in edge networks still faces several difficulties.
1) The communication between devices and the server in dynamic edge networks  may be frequently unavailable, slow, and expensive. 
2) The resources (\eg, bandwidth and battery life) are always limited in the MEC system.

These issues can be alleviated by the fact that modern edge devices usually can connect to the edge FL server via multiple communication channels (\eg, 4G, LTE and 5G).  
However, having an edge device for sending copies of local models to the FL server along multiple channels is redundant, time-consuming, and would waste resources (\eg, bandwidth, battery life and monetary cost). 


Several pioneering works have been proposed to manage system resources for efficient FL in edge networks \cite{wang2019adaptive,tran2019federated,chen2020convergence}.
However, these studies focus on reducing resource consumption, hindering performance boost in resource utilization and training efficiency.
A promising solution suggested in recent works is to incorporate gradient compression strategies into FL algorithms, which can considerably reduce the communication cost with little impact on learning outcomes \cite{stich2018sparsified,basu2019qsparse}.
However, these compression techniques are not tuned to the underlying communication channel, and may not utilize the channel resources to the fullest.

In this paper, to address the problem of how to efficiently utilize the limited resources at edge devices for optimal learning performance,
we propose a novel FL framework called layered gradient compression (LGC). 
Motivated by the layered coding techniques in video streaming, in LGC, local gradients from a device are coded into several layers and each layer is sent to the FL server along a different channel. The FL server aggregates the received layers of local gradients from devices to update the global model, and sends the result back to the devices.
We integrate gradient compression and multi-channel transmission into FL to alleviate communication and energy bottleneck.
We prove the convergence of LGC, and formally define the problem of resource-efficient FL with LGC. 
To deploy LGC in dynamic networks and resource constrained MEC systems, we then propose a learning-based algorithm for each device to dynamically adjust its local computation (\ie, the number of local stochastic descent) and communication decisions (\ie, the compression level of different layers and the layer-to-channel mapping) in each iteration.

Our \textbf{main contributions} of this paper are as follows:

\begin{itemize}
    
    \item
    To efficiently utilize the limited resources at edge devices for the optimal learning performance in dynamic edge networks, motivated by the layered coding techniques in video streaming, we propose a novel FL framework called layered gradient compression (LGC). 
    To the best of our knowledge, we are the first to propose such a layered gradient compression FL framework.

    
    \item We provide a convergence guarantee for LGC from a theoretical perspective, and formally define the problem of resource-efficient FL with LGC.
    
    
    \item We then propose a learning-based control algorithm for each device to dynamically adjust its local computation and communication decisions in each iteration, subject to dynamic edge network and resource constraints.
    
    \item We evaluate the performance of \system{} with the proposed learning-based control algorithm. Results show that using our algorithm, LGC significantly reduces the training time, improves resources utilization, while achieving a similar accuracy, compared with the baseline.
\end{itemize}

The rest of this paper is organized as follows. 
In Section \ref{sec:formulation}, we describe the framework of \system{}, prove the convergence of \system{} and define the problem of resource-efficient FL with \system{}. 
In Section 
\ref{sec:design}, we describe the design and implementation details of the learning-based control algorithm.
We show the experimental results in Section \ref{sec:evaluation}, summary the related work in Section \ref{sec:related}, and conclude this work in Section \ref{sec:conclusion}.

\section{Framework Design}
\label{sec:formulation}

This section first reviews the typical framework of FL. Then, we describe our proposed LGC mechanism and prove its convergence. Finally, we put forward the problem formulation of resource-efficient FL with LGC.

\subsection{Framework Overview}

The framework of LGC follows the typical FL pattern and consists of two parts, an edge server and $M$ devices.
In LGC, $M$ edge devices denoted by $\mathcal{M} = \{1,2,\dots,m,\dots,M\}$ collaboratively train a learning model with an edge server by iterative computations and communication.

\begin{figure}[!htbp]
\centering
\centerline{\includegraphics[scale=0.6]{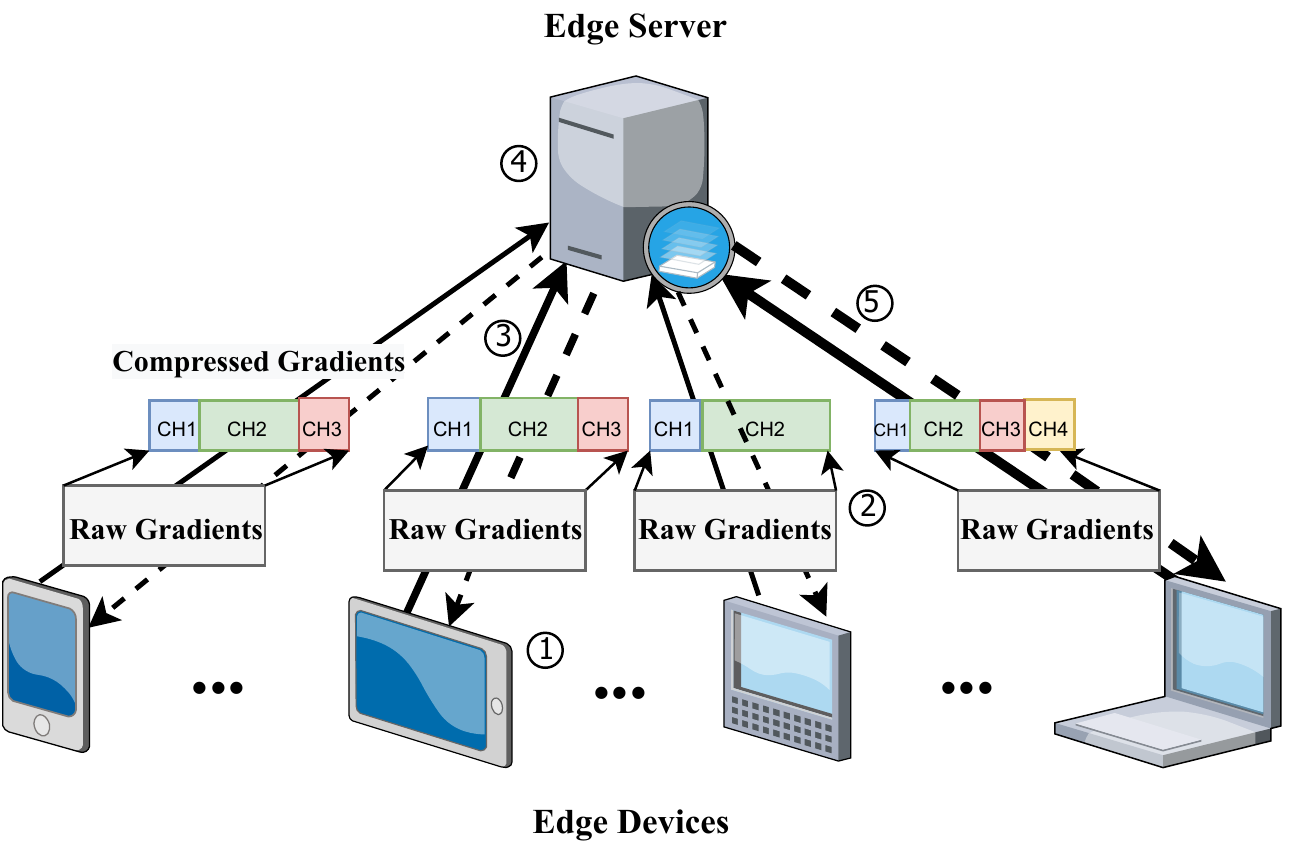}}
\caption{An overview of \system{}.}
\label{fig:architecture}
\end{figure}

To alleviate the communication bottleneck, LGC compresses the local computed gradients before transmitting and sends them through multiple channels.
Figure \ref{fig:architecture} gives and overview of \system{}.
In \system{}, each device computes the local gradients (\textcircled{1}), compress the gradients by LGC compressor (\textcircled{2}) and sends encoded layers of the compressed gradients to the edge server through multiple channels (\textcircled{3}). The server waits until the gradients from all the clients are received. It then adds them up (\textcircled{4}) and dispatches the results to all devices (\textcircled{5}). Devices then uses them to update the local model.
Multiple channels are indicated by different colors.


To compress the gradients, we consider Top$_k$ operator, an important example of sparsification operators in distributed training. And we extend it to LGC$_{\mathbf{k}}$ for 
multiple communication channels.
For any $\mathbf{x} \in \mathbb{R}^D$, Top$_k(\mathbf{x}) \in \mathbb{R}^D$ is equal to a $D$-length vector, which has at most $k$ non-zero components whose indices correspond to the indices of the largest $k$ components (in absolute value) of $\mathbf{x}$. 
Before giving the defination of LGC$_{\mathbf{k}}$,
we extend Top$_k$ compressor to Top$_{\alpha, \beta}$ ($1 \leq \alpha < \beta  \leq D$) compressor to take the sparsified top-$(\alpha, \beta)$ gradients. Specifically, for a vector $\mathbf{x} \in \mathbb{R}^D$,  Top$_{\alpha, \beta}(\mathbf{x}) \in \mathbb{R}^D$
and the $i$-th $(i = 1,2, \dots ,D)$ element of Top$_{\alpha, \beta}(\mathbf{x})$ is defined as

\begin{equation}
\text{Top}_{\alpha, \beta} (\mathbf{x}_i)=\left\{
\begin{aligned}
\mathbf{x}_i, & & \text{if \textit{thr}$_\alpha$ $\geq | x_i | >$ \textit{thr}$_\beta$}, \\
0, & & \text{otherwise},
\end{aligned}
\right.
\end{equation}
where $\mathbf{x}_i$ is the $i$-th element of $\mathbf{x}$ and \textit{thr}$_\alpha$ is the $\alpha$-th largest absolute value of the elements in $\mathbf{x}$ and and \textit{thr}$_\beta$ is the $\beta$-th largest absolute value of the elements in $\mathbf{x}$.



Modern edge devices usually can connect with multiple communication channels.  
Considering a device with $C$ channels connected to it, the traffic allocation among these channels is denoted by a vector $\mathbf{k} \in \mathbb{R}^c$.
The device codes gradient elements into different layers with Top$_{\alpha,\beta}$ compressor and gets
$\{\text{Top}_{\sum_{i=1}^{c-1} \mathbf{k}_i, \sum_{i=1}^{c} \mathbf{k}_i}(\mathbf{x})\}_{c=1}^{C}$.
Then each layer is sent to server through different channels. 
The server collects gradients from all the channels, decodes them and gets LGC$_{\mathbf{k}}(\mathbf{x})$.
For a vector $\mathbf{x} \in \mathbb{R}^D$, LGC$_{\mathbf{k}}(\mathbf{x}) \in \mathbb{R}^d$
and the $i$-th $(i = 1,2, \dots ,d)$ element of LGC$_{\mathbf{k}}(\mathbf{x})$ is defined as




\begin{equation}
\text{LGC}_{\mathbf{k}}(\mathbf{x}) = \sum_{c=1}^{C} 
\text{Top}_{\sum_{i=1}^{c} \mathbf{k}_i, \mathbf{k}_c}(\mathbf{x}).
\end{equation}



Unlike previous studies requiring an identical number of local computation and compression level across all the participants,
we propose and analyze a particular form of asynchronous operation where the devices synchronize with the master at arbitrary times. 
We also allow the participating devices to perform gradient sparsification with different compression coefficients.
This indeed helps to accommodate stragglers with poor channel conditions and thus mitigates the impacts of stale updates. 
By definition, we also allow devices to be equipped with different numbers and types of communication channels.

\begin{algorithm}[h]
 \caption{FL with Layered Gradient Compression}
  \label{sync-ef-compressed-local-sgd}
  \begin{algorithmic}[1]
    \State Initialize $\mathbf{w}^{(0)}, \bar{\bar{\mathbf{w}}}^{(0)}, \mathbf{w}^{(0)}_m, \widehat{\mathbf{w}}_{m}^{(0)}, \mathbf{e}_{m}^{(0)}, \forall m \in \mathcal{M}$.
    Suppose $\eta^{(t)}$ follows a certain learning rate schedule.
    \For{$t = 0 \mathbf{to} T-1$}
    \State \textbf{On Edge Devices:}
    \For {$m \in \mathcal{M}$ in parallel}
    \State Sampling a mini-batch $\mathcal{D}^{(t)}_{m}$ of size $b$ from $\mathcal{D}_{m}$
    \State $\widehat{\mathbf{w}}^{(t+\frac{1}{2})}_{m} \gets \widehat{\mathbf{w}}^{(t)}_{m} - \eta^{(t)} \nabla f_m(\widehat{\mathbf{w}}^{(t)}_{m}; \mathcal{D}^{(t)}_{m})$;
    \If {$t+1 \in \mathcal{I}_{m} $}
    \State $\mathbf{u}^{(t)}_{m} \gets \mathbf{e}^{(t)}_{m} + \mathbf{w}^{(t)}_{m} -
    \widehat{\mathbf{w}}^{(t+\frac{1}{2})}_{m}$
    \State $\mathbf{g}^{(t)}_{m} = \text{LGC}^{(t)}_{m}(\mathbf{u}^{(t)}_{m})$
    \State Upload $\mathbf{g}^{(t)}_{m}$ by multiple channels
    \State $\mathbf{e}^{(t+1)}_{m} \gets \mathbf{e}^{(t)}_{m} + \mathbf{w}^{(t)}_{m} - \widehat{\mathbf{w}}^{(t+\frac{1}{2})}_{m} - \mathbf{g}^{(t)}_{m}$
    \State Receive $\bar{\bar{\mathbf{w}}}^{(t+1)}$
    \State $\widehat{\mathbf{w}}^{(t+1)}_{m} \gets \bar{\bar{\mathbf{w}}}^{(t+1)}$ and ${\mathbf{w}}^{(t+1)}_{m} \gets \bar{\bar{\mathbf{w}}}^{(t+1)}$    
    \Else    
    \State $\widehat{\mathbf{w}}^{(t+1)}_{m} \gets \widehat{\mathbf{w}}^{(t+\frac{1}{2})}_{m}$
    \State $\mathbf{w}^{(t+1)}_{m} \gets \mathbf{w}^{(t)}_{m}$
    \State $\mathbf{e}^{(t+1)}_{m} \gets \mathbf{e}^{(t)}_{m}$
    \EndIf
    \EndFor
    \State \textbf{At Central Server:}
    \If {$t+1 \in \mathcal{I}_{m}, \forall m \in \mathcal{M}$}
    \State Collect $\mathbf{g}^{(t)}_{m}$, $\forall m \in \mathcal{M}$ and $\mathbf{g}^{(t)} \gets \frac{1}{M} \sum_{m=1}^{M} \mathbf{g}^{(t)}_{m}$
    \State $\bar{\bar{\mathbf{w}}}^{(t+1)} \gets \bar{\bar{\mathbf{w}}}^{(t)} - \mathbf{g}^{(t)}$ and broadcast $\bar{\bar{\mathbf{w}}}^{(t+1)}$
    \Else   
    \State $\bar{\bar{\mathbf{w}}}^{(t+1)} \gets \bar{\bar{\mathbf{w}}}^{(t)}$
    \EndIf
    \EndFor
    \State \textbf{Comment:} $\mathbf{w}^{(t+\frac{1}{2})}_{m}$ denotes an intermediate variable between iterations $t$ and $t+1$
  \end{algorithmic}
\end{algorithm}

Let $\mathcal{I}_{m} \subseteq \mathcal{T} := \{1, \dots,T\}$ with $T \in \mathcal{I}_{m}$ denote a set of indices for which device $m \in \mathcal{M}$ synchronizes with the server.
In our asynchronous setting, $\mathcal{I}_{m}$'s may be different for different devices. However, we assume that gap$(\mathcal{I}_{m}) \leq H$ holds for every $m \in \mathcal{M}$, which means that there is a uniform bound on the maximum delay in each device's update times.
Every device $m \in \mathcal{M}$ maintains a local parameter vector $\widehat{\mathbf{w}}^{(t)}_{m}$ which is updated in each iteration $t$. If $t \in \mathcal{I}_m$, the error-compensated update $\mathbf{g}^{(t)}_{m}$ computed on the net progress made since the last synchronization is sent to the server with multi-channel communication, and updates its local memory $\mathbf{e}^{(t)}_{m}$. Upon receiving $\mathbf{g}^{(t)}_{m}$ from every device $m \in \mathcal{M}$ which sent its gradients, master aggregates them, updates the global parameter vector, and sends the new model $\mathbf{w}^{(t+1)}$ to all the workers; upon receiving which, they set their local parameter vector $\widehat{\mathbf{w}}^{(t+1)}_{m}$ to be equal to the global parameter vector $\mathbf{w}^{(t+1)}$. 
Our algorithm is summarized in Algorithm \ref{sync-ef-compressed-local-sgd}.

\subsection{Convergence Analysis}

We consider the following two standard assumptions on the local loss functions $f_{m}: \mathbb{R}^{d} \rightarrow \mathbb{R}, \forall m \in \mathcal{M}$

\begin{assumption}
(Smoothness): $f_{m}(\cdot)$ is $L$-smooth, i.e., for every $\mathbf{w}, \mathbf{w}^{\prime} \in \mathbb{R}^{d}$, we have

\begin{equation}
f_{m}(\mathbf{w}) \leq f_{m}\left(\mathbf{w}^{\prime}\right)+<\nabla f_{m}(\mathbf{w}), \mathbf{w}^{\prime}-\mathbf{w}>+\frac{L}{2}\left\|\mathbf{w}^{\prime}-\mathbf{w}\right\|^{2}.
\end{equation}
\end{assumption}

\begin{assumption}
(Bounded variances and second momentum):
For every $\mathbf{w}_{m}^{(t)} \in \mathbb{R}^{d}$ and $t \in \mathbb{Z}^{+}$, there exists constants $\sigma>0$ and $G \geq \sigma$ such that:

\begin{subequations}
\begin{equation}
\mathbb{E}_{\mathcal{D}_{m}^{(t)} \subset \mathcal{D}_{m}} \left[\| \nabla f_{m}\left(\mathbf{w}_{m}^{(t)} ; \mathcal{D}_{m}^{(t)}\right)-\nabla f_{m}\left(\mathbf{w}_{m}^{(t)}\right)\|^{2}\right] \leq \sigma^{2}, \forall m,
\end{equation}
\begin{equation}
\mathbb{E}_{\mathcal{D}_{m}^{(t)} \subset \mathcal{D}_{m}}\left[\left\|\nabla f_{m}\left(\mathbf{w}_{m}^{(t)} ; \mathcal{D}_{m}^{(t)}\right)\right\|^{2}\right] \leq G^{2}, \forall m.
\end{equation}
\end{subequations}

\end{assumption}

To state our results, we need the following definition from \cite{stich2018local}.

\begin{definition}
(Gap). Let 
$\mathcal{I} = \{ t_0, t_1, \dots, t_k$ \}, where $t_i < t_{i+1}$ for $i = 0,1, \dots, k - 1$. 
The gap of $\mathcal{I}$ is defined as $gap(\mathcal{I}) := \max_{i \in \{1,\dots,k\}} {(t_i - t_{i-1})}$, which is equal to the maximum difference between any two consecutive synchronization indices.
\end{definition}

We extent Lemma 4 in \cite{basu2019qsparse} and get the following lemma.

\begin{lemma}
\label{lemma_memory_contraction}
(Memory contraction). Let $\operatorname{gap}\left(\mathcal{I}_m\right) \leq H, \forall m \in \mathcal{M}$ and $\eta^{(t)}=\frac{\xi}{a+t}$, where $\xi$ is a constant and $a>\frac{4 H}{\gamma}$. Then there exists a constant $C \geq \frac{4 a \gamma_{m}\left(1-\gamma_{m}^{2}\right)}{a \gamma_{m}-4 H}$, the following holds for every $t \in \mathbb{Z}^{+}$ and $m \in \mathcal{M}$:

\begin{equation}
\mathbb{E}\left\|\mathbf{e}^{(t)}_{m}\right\|_{2}^{2} \leq 4 \frac{(\eta^{(t)})^{2}}{\gamma_{m}^{2}} C H^{2} G^{2}.
\end{equation}
\end{lemma}



We leverage the perturbed iterate analysis as in \cite{mania2015perturbed,stich2018sparsified} to provide convergence guarantees for \system{}. Under the above assumptions, the following theorems hold for Algorithm \ref{sync-ef-compressed-local-sgd}.

\begin{theorem}
\label{theorem_async_convergence_rate}
(Smooth and strongly convex case with a decaying learning rate).
Let $f_{m}(\mathbf{w})$ be $L$-smooth and $\mu$-strongly convex, $\forall m \in \mathcal{M}$. 
Let $\{\widehat{\mathbf{w}}^{(t)}_{m}\}_{t=0}^{T-1}$ be generated according to Algorithm \ref{sync-ef-compressed-local-sgd} with $\mathcal{C}_m^{(t)}$, for step sizes $\eta^{(t)} = 8/\mu(a+t)$ with $gap(\mathcal{I}) \leq H$, where $a > 1$ is such that we have $a > \max \{ 4H/\gamma, 32 \kappa, H \}$, $\kappa = L/\mu$. The following holds

\begin{equation}
\begin{aligned}
& \mathbb{E} [f (\overline{\mathbf{w}}^{(T)})] - f^{*} \\
& \leq
\frac{La^3}{4S} \| \mathbf{w}^{(0)} - \mathbf{w}^{*} \|_2^2 + \frac{8LT(T+2a)}{\mu^2 S} A + \frac{128 LT}{\mu^3 S}B,
\end{aligned}
\end{equation}
where


\begin{subequations}

\begin{equation}
C = \min_{m\in\mathcal{M}} \frac{4a \gamma_m (1-\gamma_m^2)}{a\gamma_m - 4H},
\end{equation}
\begin{equation}
C_1 = \frac{192}{M}\sum_{m=1}^{M} (4-2\gamma_m) (1+\frac{C}{\gamma_m^2}),
\end{equation}
\begin{equation}
C_2 =  \frac{8}{M} \sum_{m=1}^{M}(4-2\gamma_m)(1 + \frac{C}{\gamma_m^2}),
\end{equation}
\begin{equation}
A  = \frac{\sum_{m=1}^M \sigma_m^2}{bM^2},
\end{equation}
\begin{equation}
\begin{aligned}
B & = (\frac{3\mu}{2} + 3L) (\frac{12CG^2 H^2}{\gamma^2} + C_{1} (\eta^{(t)})^2 H^4G^2) \\
& + 24(1+C_{2}H^2) LG^2H^2,  \qquad\qquad \qquad\qquad \qquad
\end{aligned}
\end{equation}
\begin{equation}
\overline{\mathbf{w}}^{(T)} 
 = \frac{1}{S} \sum_{t=0}^{T-1} \left[ s^{(t)} \left( \frac{1}{M} \sum_{m=1}^M \widehat{\mathbf{w}}^{(t)}_{m} \right) \right] = \frac{1}{S} \sum_{t=0}^{T-1} s^{(t)} \widehat{\mathbf{w}}^{(t)},
\end{equation}
\begin{equation}
s^{(t)}  = (a+t)^2,
\end{equation}
\begin{equation}
S  = \sum_{t=0}^{T-1} s^{(t)} \geq \frac{T^3}{3}.
\end{equation}
\end{subequations}



\end{theorem}

\begin{corollary}
For gap$(\mathcal{I}_m) \leq H$, 
$a > \max \{ 4H/\gamma, 32 \kappa, H \}$, $\sigma_{\textit{max}} = \max_{m\in \mathcal{M}} \sigma_{m}$, 
if $\{ \mathbf{w}_m^{(t)}\}_{t=0}^{T-1} $ is generated according to Algorithm \ref{sync-ef-compressed-local-sgd} 
and using $\mathbb{E}\| \mathbf{w}^{(0)} - \mathbf{w}^{*} \|_2^2 \leq \frac{4G^2}{\mu^2}$ from Lemma 2 in \cite{rakhlin2011making}, we have

\begin{equation}
\begin{aligned}
\mathbb{E} [f (\overline{\mathbf{x}}^{(T)})] - f^{*}
\leq
& \mathcal{O} \left( \frac{G^2H^3}{\mu^2 \gamma^3 T^3} \right) \\
+ & \mathcal{O} \left( \frac{\sigma_{\textit{max}}^2}{\mu^2 b RT} + \frac{H \sigma_{\textit{max}}^2}{\mu^2 bR\gamma T^2} \right)
\\
+ & \mathcal{O} \left( \frac{G^2}{\mu^3 \gamma^2T^2} (H^2 + H^4) \right).
\end{aligned}
\end{equation}
\end{corollary}

\subsection{Problem Formulation}

In this part, we define resource-efficient FL with LGC.
Considering the resources of different mobile devices varies, we formulate the optimization problem to minimize global loss function under resource constraints as follows.




\begin{equation}
\min_{\{ T, H_m^{(t)}, D_{m, n}^{(t)} \}}
f({\mathbf{w}}^{(T)}),
\end{equation}
subject to,

\begin{subequations}

\begin{equation}
\begin{aligned}
\sum_{t=1}^T \left( E_{m, r, \textit{comp}}^{(t)} H_{m}^{(t)}  + \sum_{n=1}^{N} 
E_{m, r, \textit{comm}}^{(t)}
D_{m, n}^{(t)} \right)
\leq B_{m, r} &, \\    
\forall m \in \mathcal{M}, \forall r \in \mathcal{R} &
\end{aligned}
\end{equation}
\begin{equation}
 \sum_{n=1}^{N} D_{m, n}^{(t)} \leq D ,
\forall m \in \mathcal{M}, \forall t \in \mathcal{T},
\end{equation}
\begin{equation}
 H_{m}^{(t)} \leq H, 
\forall m \in \mathcal{M}, \forall t \in \mathcal{T},
\end{equation}

\end{subequations}

where $E_{m, r, \textit{comp}}^{(t)}$ is the total resource consumption for local computation of device $m$ for resource $r$ in round $t$ and $E_{m, r, \textit{comm}}^{(t)}$ is the resource consumption factor for communication of device $m$ for resource $r$ in round $t$. $H_m^{(t)}$ represents the number of local update steps at device $m$ in round $t$. $D_{m,n}^{(t)}$ indicates the traffic allocation for channel $n$ at device $m$ in round $t$. $B_{m, r}$ represents the total budget for resource $r$ in device $m$.

Since FL is typically deployed in highly dynamic edge networks, a learning-based method could be useful to adaptively adjust the local computation and communication decision, while satisfying the resource constraints at each epoch in MEC. 
\section{Control Algorithm Design}
\label{sec:design}



In this section, we propose a learning-based control algorithm for \system{} to achieve resource-efficient FL. We first introduce the workflow of the deep reinforcement learning (DRL) algorithm and then describe how to transform the formulated problem into a DRL procession. 

\subsection{Deep Reinforcement Learning Mechanism}

Different from some traditional approaches using predefined rules or model-based heuristics, the DRL based method aims to learn a general action set based on the current system state and the given reward.
This is critical for deploying \system{} in a highly dynamic environment.

\begin{figure}[ht]
\centering
\centerline{\includegraphics[scale=0.4]{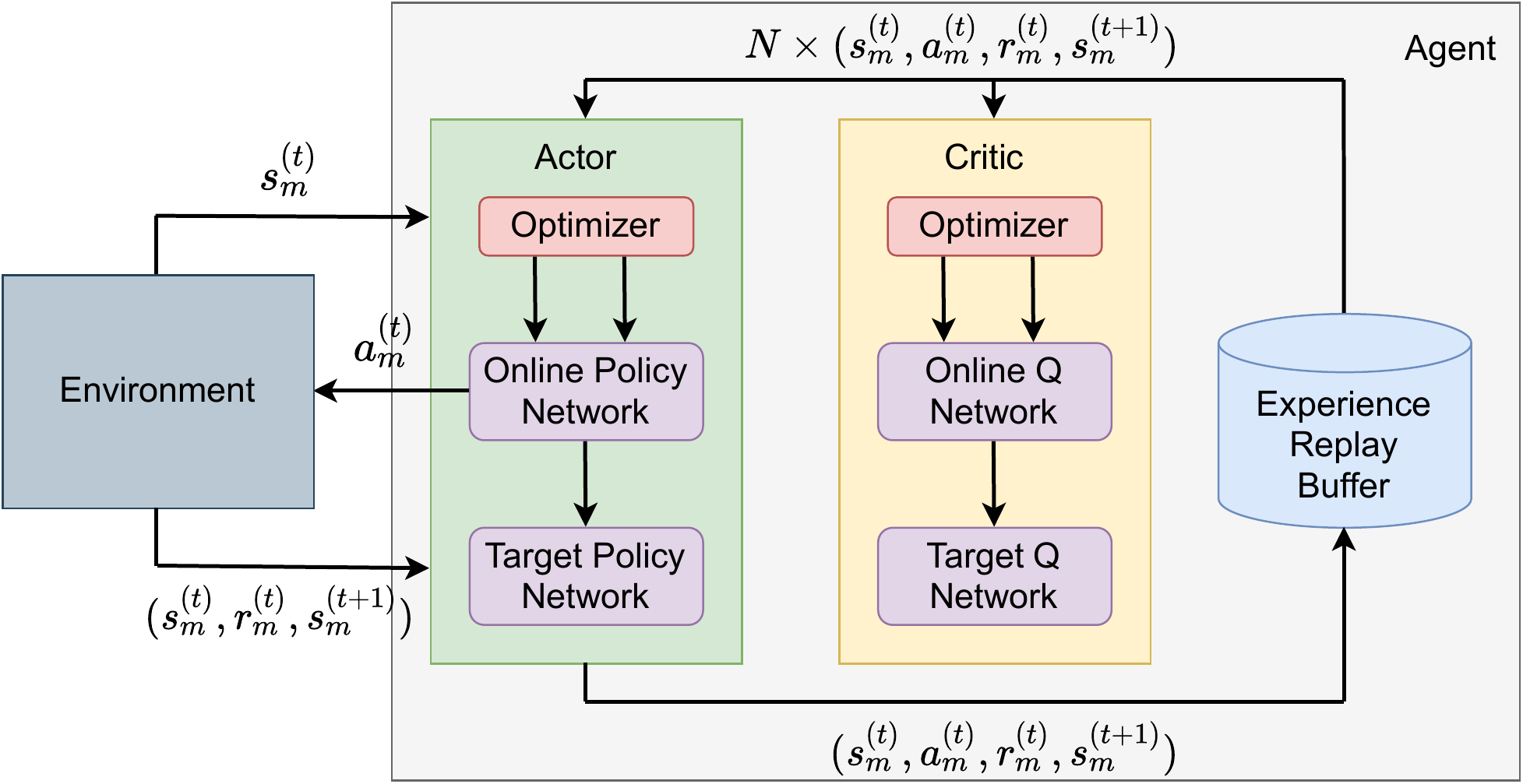}}
\caption{The workflow of the DRL algorithm.}
\label{drl}
\end{figure}

The workflow of the DRL method is illustrated in Figure~\ref{drl}.
At each epoch $t$, for each device $m$, it measures its state $\mathbf{s}_m^{(t)}$, computes the corresponding reward ${r}_m^{(t)}$, and chooses its action $a_m^{(t)}$ based on its policy $\pi_m^{(t)}$. After device $m$ updates its state to $\mathbf{s}_m^{(t+1)}$ at the next epoch $t+1$, it puts the tuple $(\mathbf{s}_m^{(t)}, \mathbf{a}_m^{(t)}, {r}_m^{(t)}, \mathbf{s}_m^{(t+1)})$ in a replay buffer for experience accumulation.
A critic network then reads from the replay buffer and updates the policy to $\pi_m^{(t+1)}$ together with the optimizer. In particular, $\pi_m^{(t+1)}$ is updated with the goal of maximizing the
accumulative rewards $\mathbf{R}_m^{(t)}=\sum\limits_{t=0}^{\infty} \gamma_m^{(t)} r_m^{(t)}$,
where $\gamma \in (0,1]$ is a discount factor of future
rewards.

\subsection{Model Design}

To implement the formulated problem using DRL techniques, we first specify the state space, the action space and the reward function as below.

\para{State Space.} 
The state of each agent contains the current resource consumption of each type of resource.
We denote the state space $\mathcal{S}_m =  \{ \mathbf{s}_m^{(t)}, \forall t\in \mathcal{T} \}$.
And we define $s_m^{(t)}$ as follows

\begin{equation}
s_m^{(t)} = (
\mathbf{E}_{m, \text{comm}}^{(t)}, \mathbf{E}_{m, \text{comp}}^{(t)})
,
\end{equation}

where

\begin{subequations}
\begin{equation}
\mathbf{E}_{m, \text{comm}}^{(t)}  = (
E_{m, 1, \text{comm}}^{(t)},
\cdots,
E_{m, r, \text{comm}}^{(t)},
\cdots,
E_{m, R, \text{comm}}^{(t)}
),
\end{equation}
\begin{equation}
\mathbf{E}_{m, \text{comp}}^{(t)} = (
E_{m, 1, \text{comp}}^{(t)},
\cdots,
E_{m, r, \text{comp}}^{(t)},
\cdots,
E_{m, R, \text{comp}}^{(t)}
).
\end{equation}
\end{subequations}


The state variables are described as follows.

\begin{itemize}
    \item $E_{m, r, \text{comm}}^{(t)}$ represents consumption factor for communication of resource $r$ at device $m$ in round $t$.
    \item $E_{m, r, \text{comp}}^{(t)}$ represents total consumption for local computation of resource $r$ at device $m$ in round $t$.
\end{itemize}

\para{Action Space.}
Each device $m$ has an action space denoted as $\mathcal{A}_m = \left \{ \mathbf{a}_m^{(t)}, \forall t \in \mathcal{T} \right \}$.
On receiving state $s_m^{(t)}$, the agent $m$ needs to choose its local computation
and communication decisions
.
Specifically, an action can be represented as 



\begin{equation}
\begin{aligned}
a_m^{(t)} = (H_m^{(t)}, \mathbf{D}_m^{(t)}),
\end{aligned}
\end{equation}
where $\mathbf{D}_m^{(t)} = (D_{m,1}^{(t)}, \cdots, D_{m,n}^{(t)}, \cdots, D_{m,N}^{(t)})$. 


The action variables are described as follows.

\begin{itemize}
    \item $H_m^{(t)}$ represents the number of local iterations at device $m$ in round $t$.
    \item $D_{m,n}^{(t)}$ represents the number of gradient entries sent through channel $n$ at device $m$ in round $t$.
\end{itemize}




\para{Reward Function.}
At each training epoch $t$, the agent $m$ will get a reward $r(s_m^{(t)};a_m^{(t)};s_m^{(t+1)})$ under a certain state $s_m^{(t)}$ after executing action $a_m^{(t)}$. 
The objective function of this work is to minimize the global loss function $\varepsilon^{(T)} = \sum_{m=1}^{M}\varepsilon_{m}^{(T)}$ under resource constraints. Hence, we minimize $\varepsilon_{m}^{(T)}$ for each device $m$ under its resource constraints.
We first define the utility function over resource $r$ at device $m$ in iteration $t$ as follows:

\begin{equation}
U^{(t)}_{m,r} = \frac{\delta^{(t)}}{\epsilon^{(t)}_{m,r}},
\end{equation}
where

\begin{subequations}
\begin{equation}
\delta^{(t)}_m = 
\varepsilon_m^{(t)} - \varepsilon_m^{(t-1)},
\end{equation}
\begin{equation}
\epsilon^{(t)}_{m,r} =  \left( E_{m, r, \textit{comp}}^{(t)} H_{m}^{(t)}  + \sum_{n=1}^{N} E_{m, r, \textit{comm}}^{(t)} D_{m, n}^{(t)} \right).
\end{equation}
\end{subequations}

Then we define the reward function as the weighted averaging utility function over $R$ types of resources at device $m$ in iteration $t$ as follows:

\begin{equation}
\begin{aligned}
r(s^{(t)}_m;a^{(t)}_m;s^{(t+1)}_m)
&= \sum_{r=1}^{R} \alpha_{r} \frac{U_{m, r}^{(t+1)}}{U_{m, r}^{(t)}},
\end{aligned}
\end{equation}
where $\alpha_{r}$ is the weight of utility function $U_{m,r}^{(t)}$.


\subsection{DRL Algorithm Details}

In our framework, each device dynamically decides its number of local iterations, gradient compression ratio and traffic allocation among different channels based on the state-of-the-art Deep Deterministic Policy Gradient (DDPG) algorithm \cite{lillicrap2015continuous}. Specifically, the algorithm maintains a parameterized critic function and actor function. As shown in Fig.~\ref{drl}, the critic function $Q(\mathbf{s}_m^{(t)},\mathbf{a}_m^{(t)}|\theta_m^{Q})$ is implemented by a Deep Q-Network (DQN) where $\theta_m^Q$ denotes the weight vector of DQN. The actor function  $\pi(\mathbf{s}_m^{(t)}|\theta_m^\pi)$ is implemented by DNN where $\theta_m^{\pi}$ is the weight vector of the DNN. If the agent under state $\mathbf{s}_m^{(t)}$ take an action $\mathbf{a}_m^{(t)}$ at epoch $t$, the $Q$ value of the critic function will be returned as follows

\begin{equation}
Q(\mathbf{s}_m^{(t)}, \mathbf{a}_m^{(t)})=\mathbb{E}\left [ \mathbf{R}_m|\mathbf{s}_m^{(t)}, \mathbf{a}_m^{(t)} \right ],
\end{equation}
where $\mathbf{R}_m= \sum\limits_{k=t}^T \gamma_m^{(t)} r ( \mathbf{s}_m^{(t)}, \mathbf{a}_m^{(t)} )$. Let $y_m^{(t)}$ be the target value at epoch $t$. It can be evaluated as

\begin{equation}
\label{eq:target_value}
\begin{aligned}
& y_m^{(t)}=r ( \mathbf{s}_m^{(t)},\mathbf{a}_m^{(t)} ) &\\
& \quad \quad+ \gamma_m^{(t)} Q(\mathbf{s}_m^{(t+1)}, \pi(\mathbf{s}_m^{(t+1)}| \theta_m^{\pi})| \theta_m^{Q} ),
\end{aligned}
\end{equation}
where $\gamma_m^{(t)}$ denotes the discount factor for future rewards at edge device $m$ at epoch $t$.

\section{Evaluation}
\label{sec:evaluation}
We describe the implementation of \system{} and verify its performance in this section. 
We first clarify our environment settings, and then show the experimental results.

\subsection{Experiment Settings}
\label{subsec_settings}

\para{{Baselines.}}
\label{subsubsec_baselines}
To illustrate the effectiveness of \system{}, 
we implement \system{} with a learning-based resource-efficient control algorithm, and we compare them with the following baseline FL mechanisms.  
\begin{itemize}
	\item \textbf{FedAvg} \cite{mcmahan2017communication} performs a fixed number of local computation in each round and aggregates the models in a centralized and synchronous paradigm. 
	\item \textbf{LGC without DRL} performs fixed number of local computations and makes the same communication decisions for each round. We use this as a baseline to show the benefits of the learning-based control algorithm.
\end{itemize}

\para{{Datasets and Models.}}
\label{subsub_datasets_models}
The experiments are conducted over three different models
(\ie., LR, CNN and RNN which are implemented by open source FedML framework \cite{he2020fedml})
and two real datasets (\ie, MNIST and Shakespeare). 

\begin{itemize}
    \item LR \cite{gortmaker1994theory} and CNN \cite{albawi2017understanding} are trained over MNIST \cite{lecun1998gradient}, which is composed of 60,000 handwritten digits for training and 10,000 for testing.
    \item RNN  is trained over Shakespeare. Shakespeare includes 40,000 lines from a variety of Shakespeare's plays.
\end{itemize}

\para{Performance Metrics.}
\label{subsubsec_performance_metrices}
In our experiments, we mainly adopt the following metrics to evaluate the performance of our proposed framework.
\begin{itemize}
    \item \textbf{Training loss} measures the difference between the predicted values and the actual values.
    The performance of both the DRL and the training models are all evaluated.
    \item \textbf{Reward} of DRL is the return of the reward function during one DRL training episode.
    \item \textbf{Model accuracy} is the proportion of correctly classified samples to all samples in the dataset. 
    \item \textbf{Energy consumption} caused by local computation and communication, which indicates the battery usage.
    \item \textbf{Money cost} denotes the money spent for the  training procedure.
\end{itemize}
\begin{figure*}[h]
\centering
    \subfigure[Loss vs. epochs]{
        \label{fig:lr_loss}
        \begin{minipage}[t]{0.23\linewidth}
            \centering
            \includegraphics[width=1.8in]{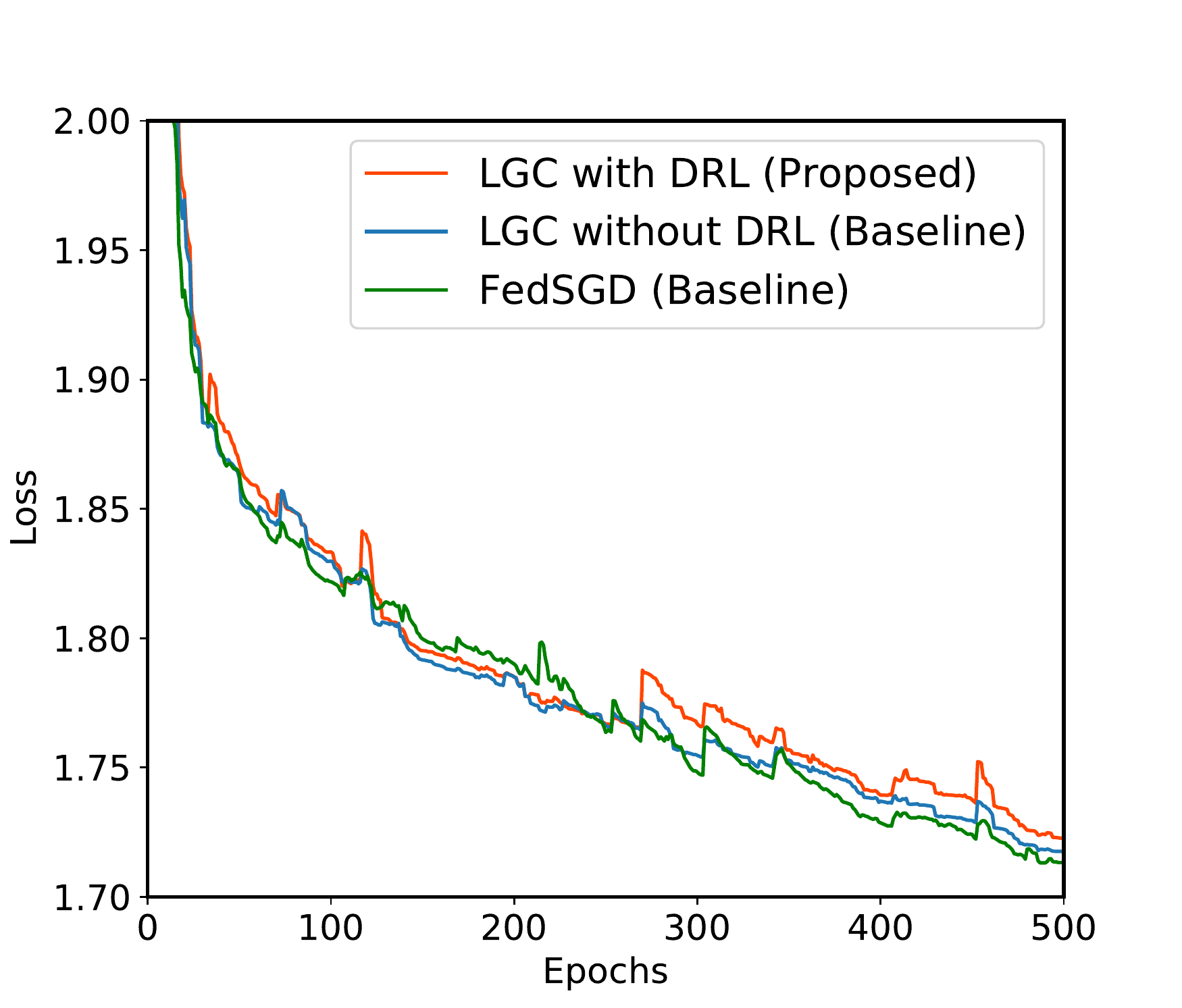}
        \end{minipage}
    }
    \subfigure[Accuracy vs. epochs]{
        \label{fig:lr_acc}
        \begin{minipage}[t]{0.23\linewidth}
            \centering
            \includegraphics[width=1.8in]{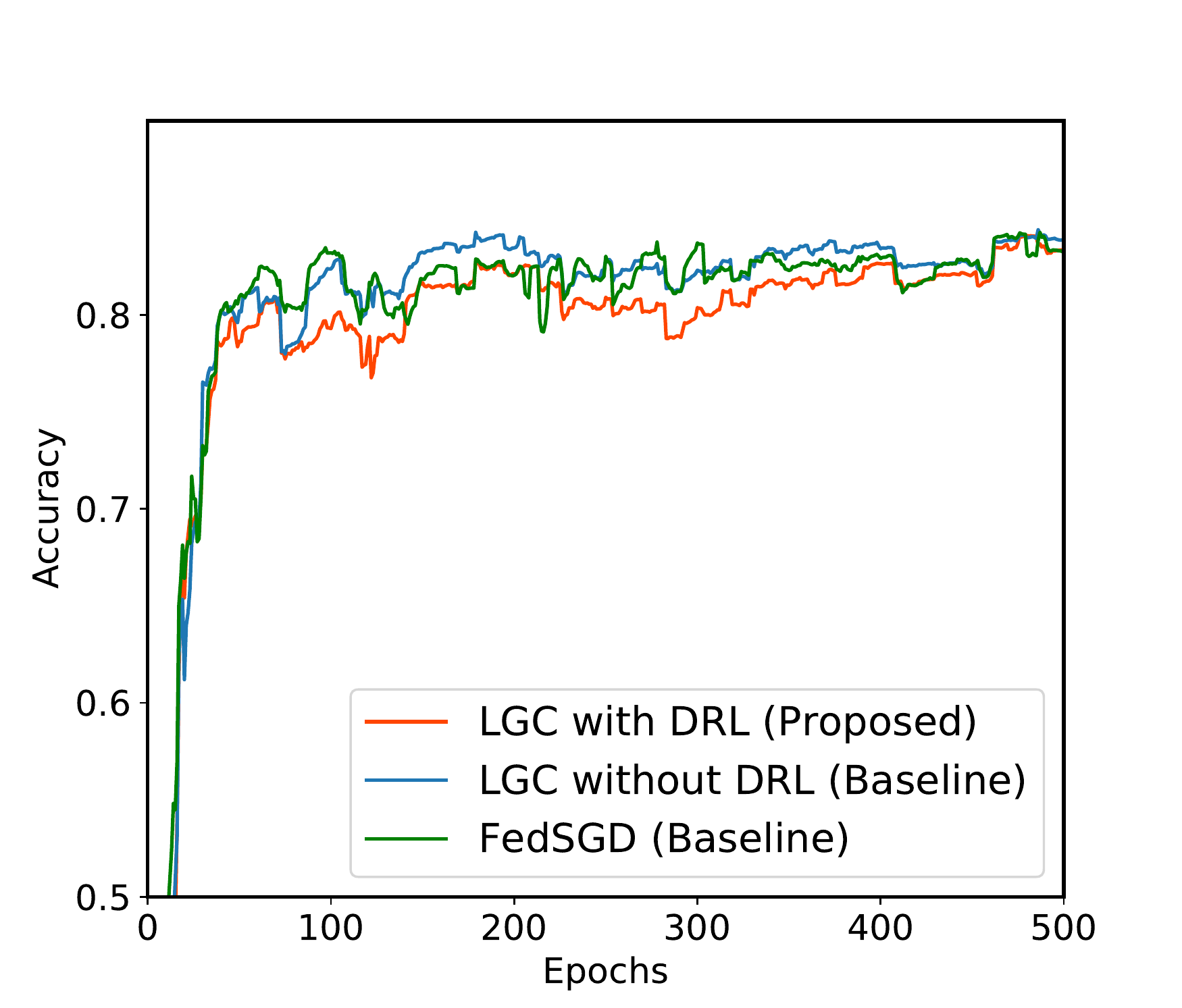}
        \end{minipage}
    }
    \subfigure[Accuracy vs. energy consumption]{
        \label{fig:lr_energy}
        \begin{minipage}[t]{0.23\linewidth}
            \centering
            \includegraphics[width=1.8in]{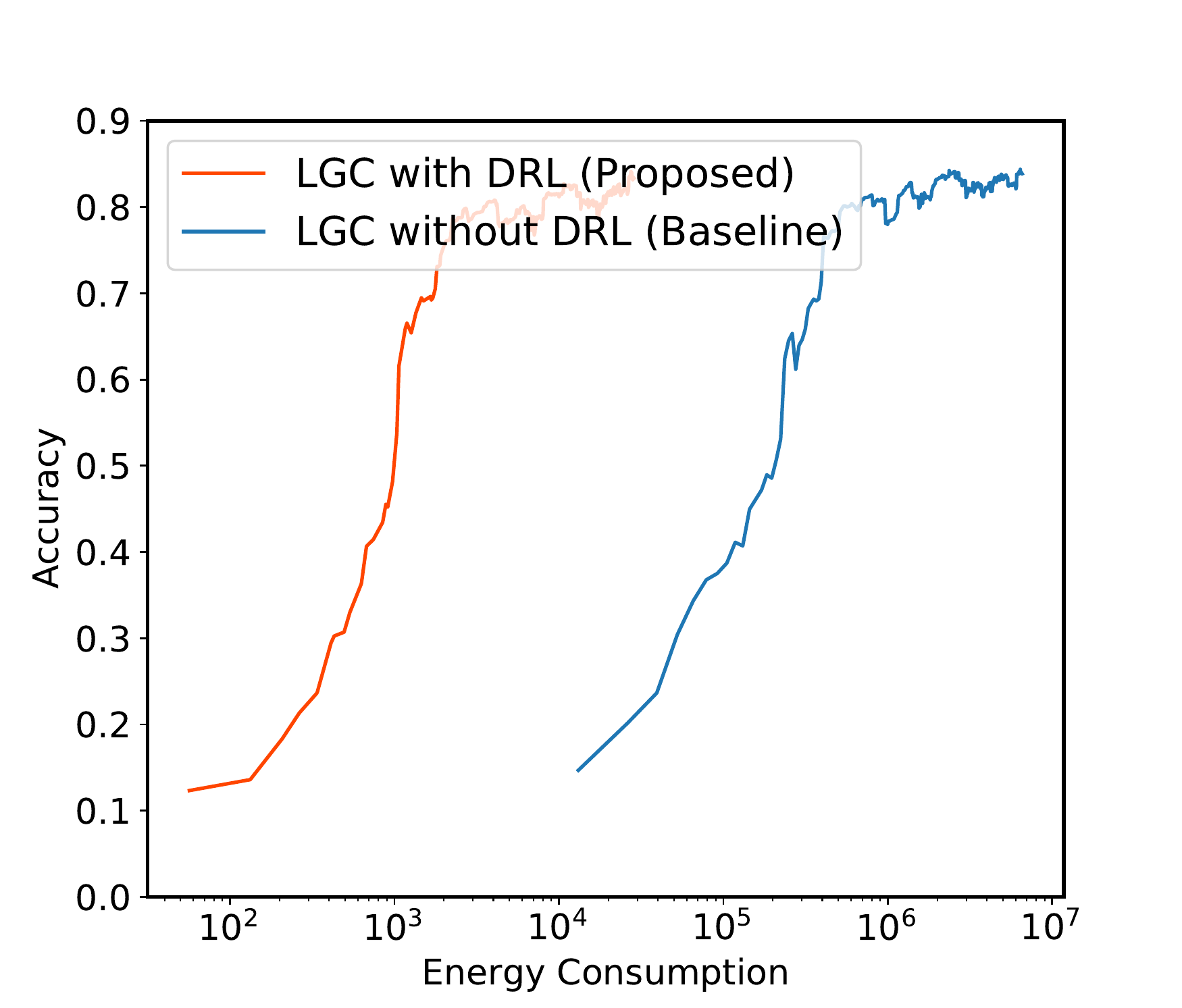}
        \end{minipage}
    }    
    \subfigure[Accuracy vs. money cost]{
        \label{fig:lr_money}
        \begin{minipage}[t]{0.23\linewidth}
            \centering
            \includegraphics[width=1.8in]{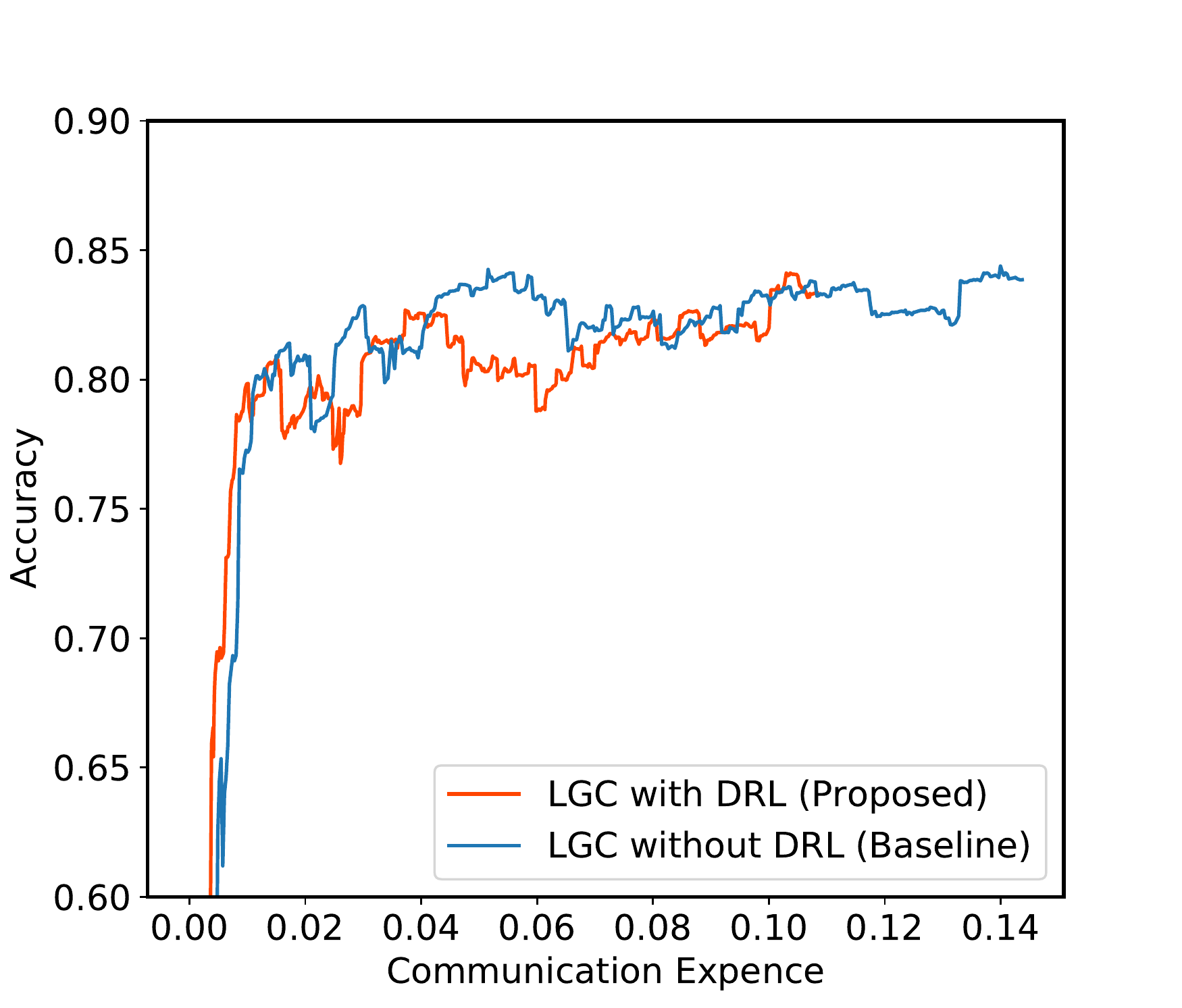}
        \end{minipage}
    }
\caption{Convergence curves of different mechanisms with LR on MNIST.}
\label{fig:lr_on_mnist}
\end{figure*}

\begin{figure*}[h]
\centering
    \subfigure[Loss vs. epochs]{
        \label{fig:reward}
        \begin{minipage}[t]{0.23\linewidth}
            \centering
            \includegraphics[width=1.8in]{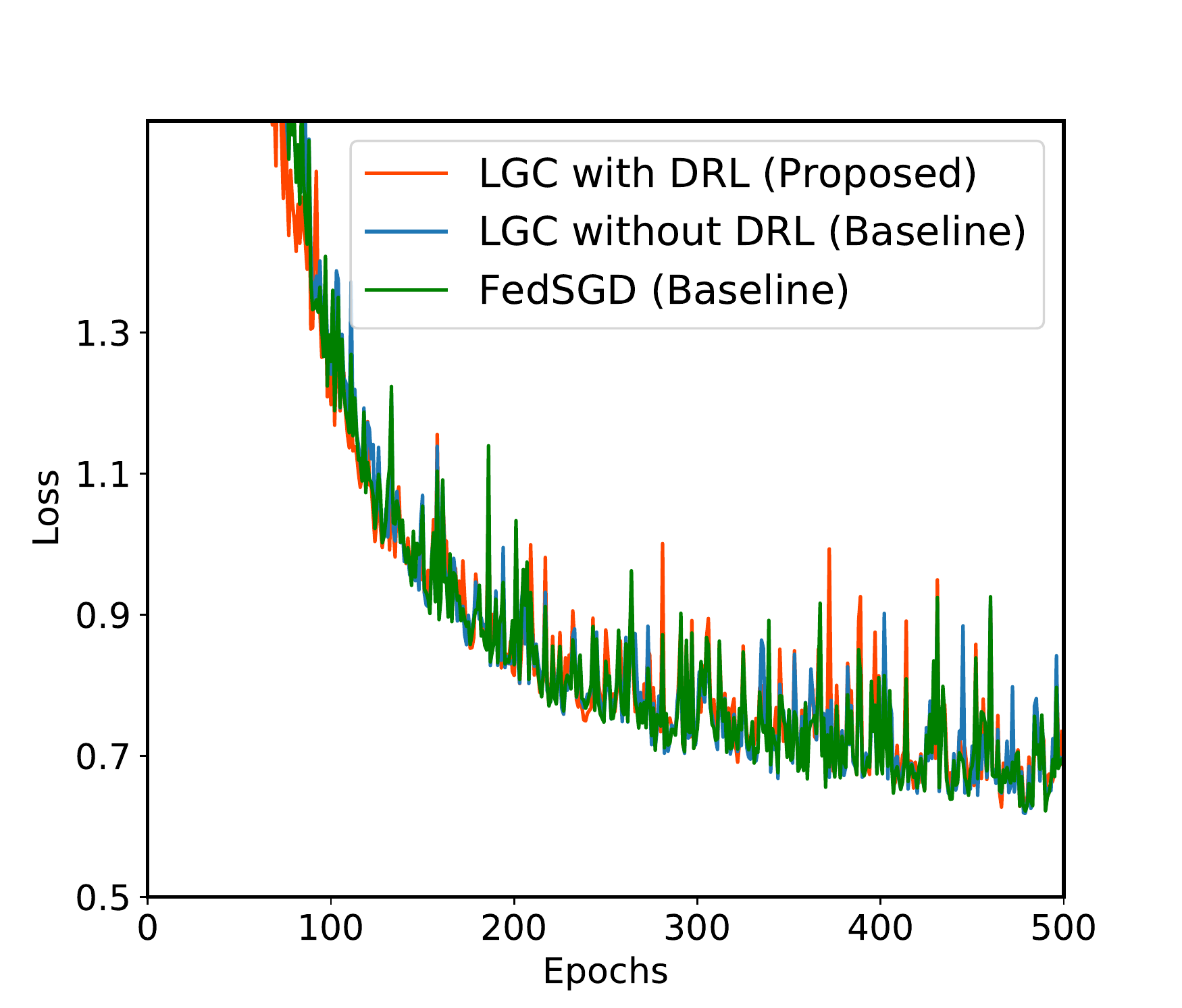}
        \end{minipage}
    }
    \subfigure[Accuracy vs. epochs]{
        \label{fig:mnist_lr}
        \begin{minipage}[t]{0.23\linewidth}
            \centering
            \includegraphics[width=1.8in]{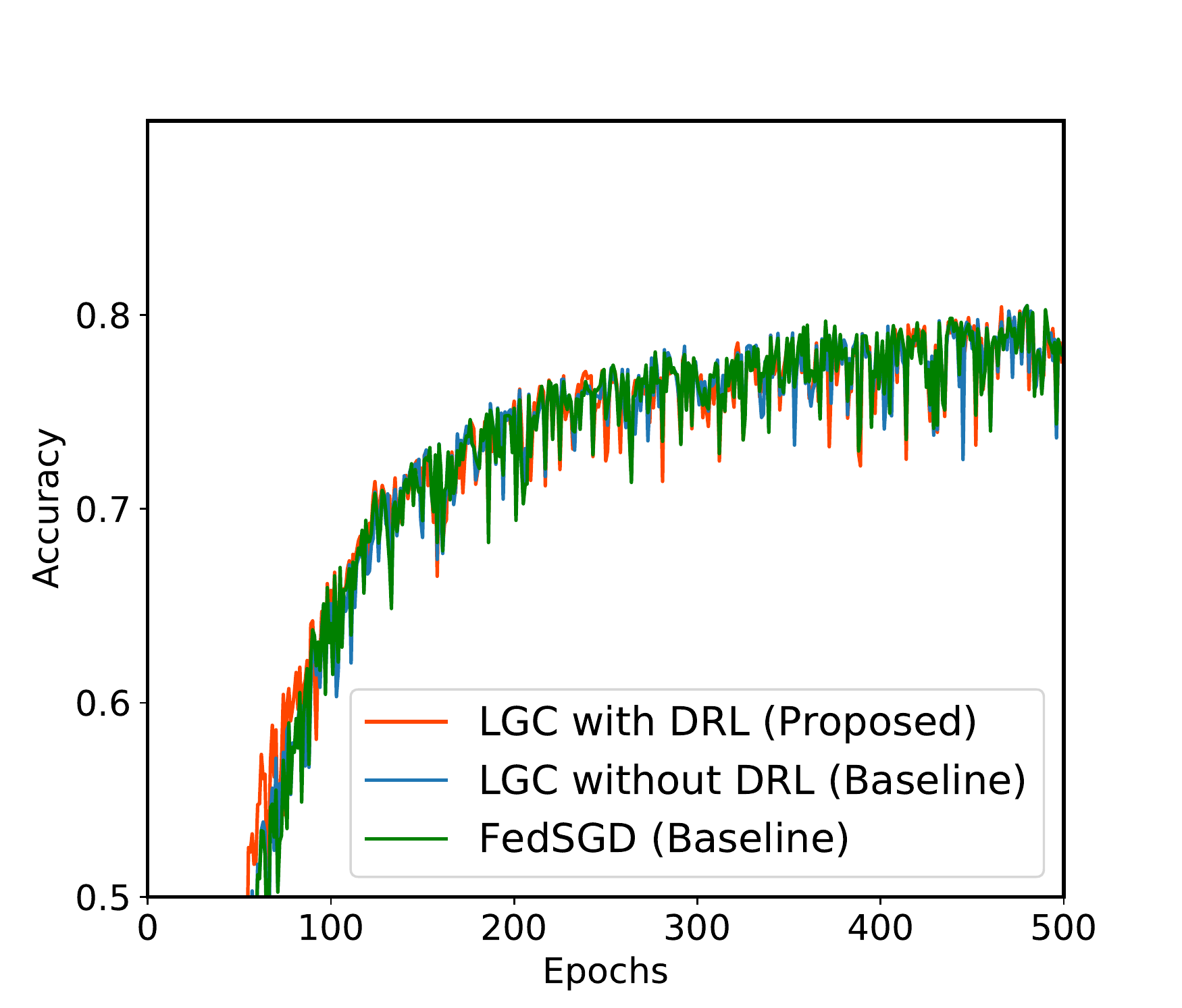}
        \end{minipage}
    }
    \subfigure[Accuracy vs. energy consumption]{
        \label{fig:cifar10_lenet}
        \begin{minipage}[t]{0.23\linewidth}
            \centering
            \includegraphics[width=1.8in]{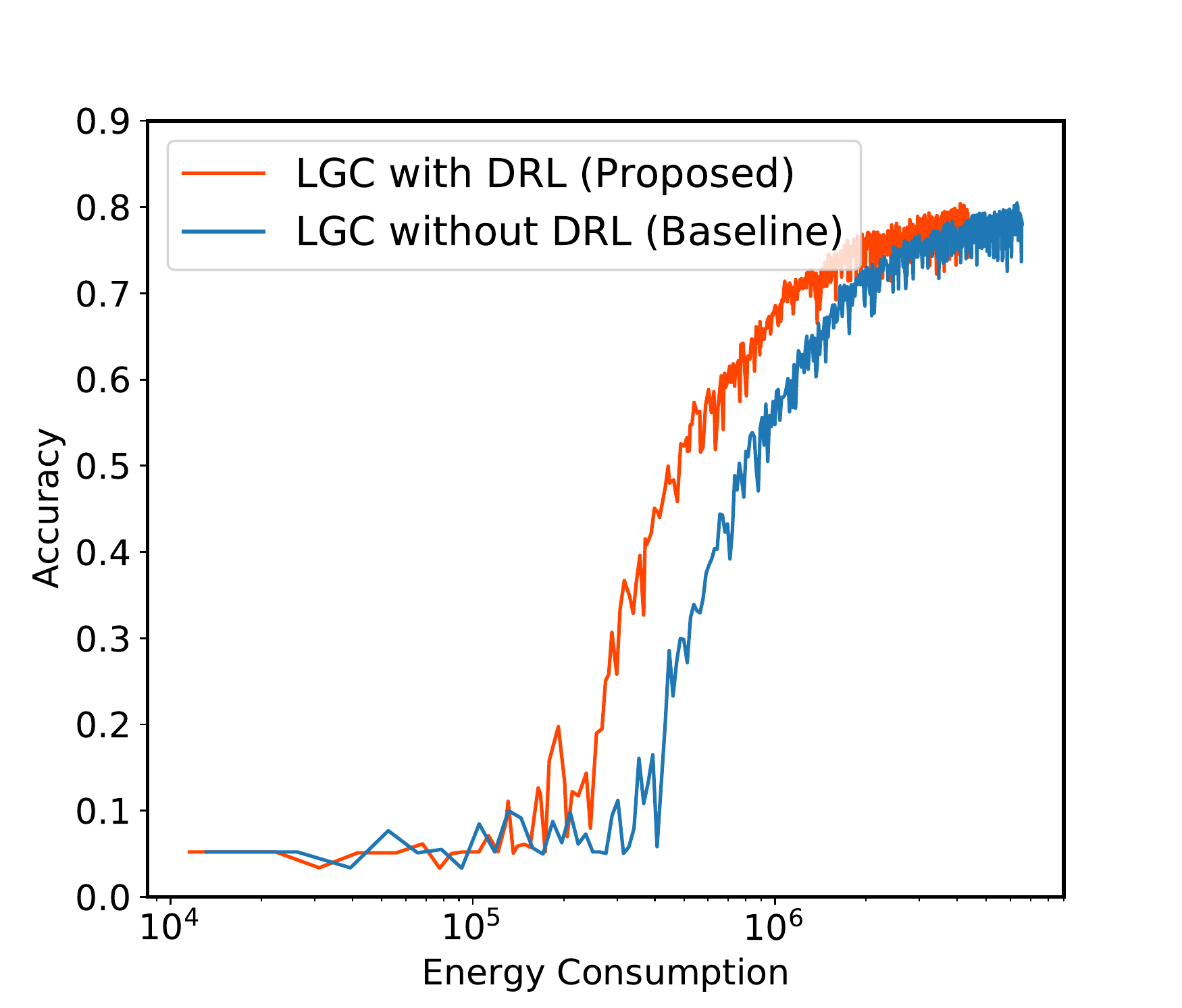}
        \end{minipage}
    }    
    \subfigure[Accuracy vs. money cost]{
        \label{fig:mnist_lenet}
        \begin{minipage}[t]{0.23\linewidth}
            \centering
            \includegraphics[width=1.8in]{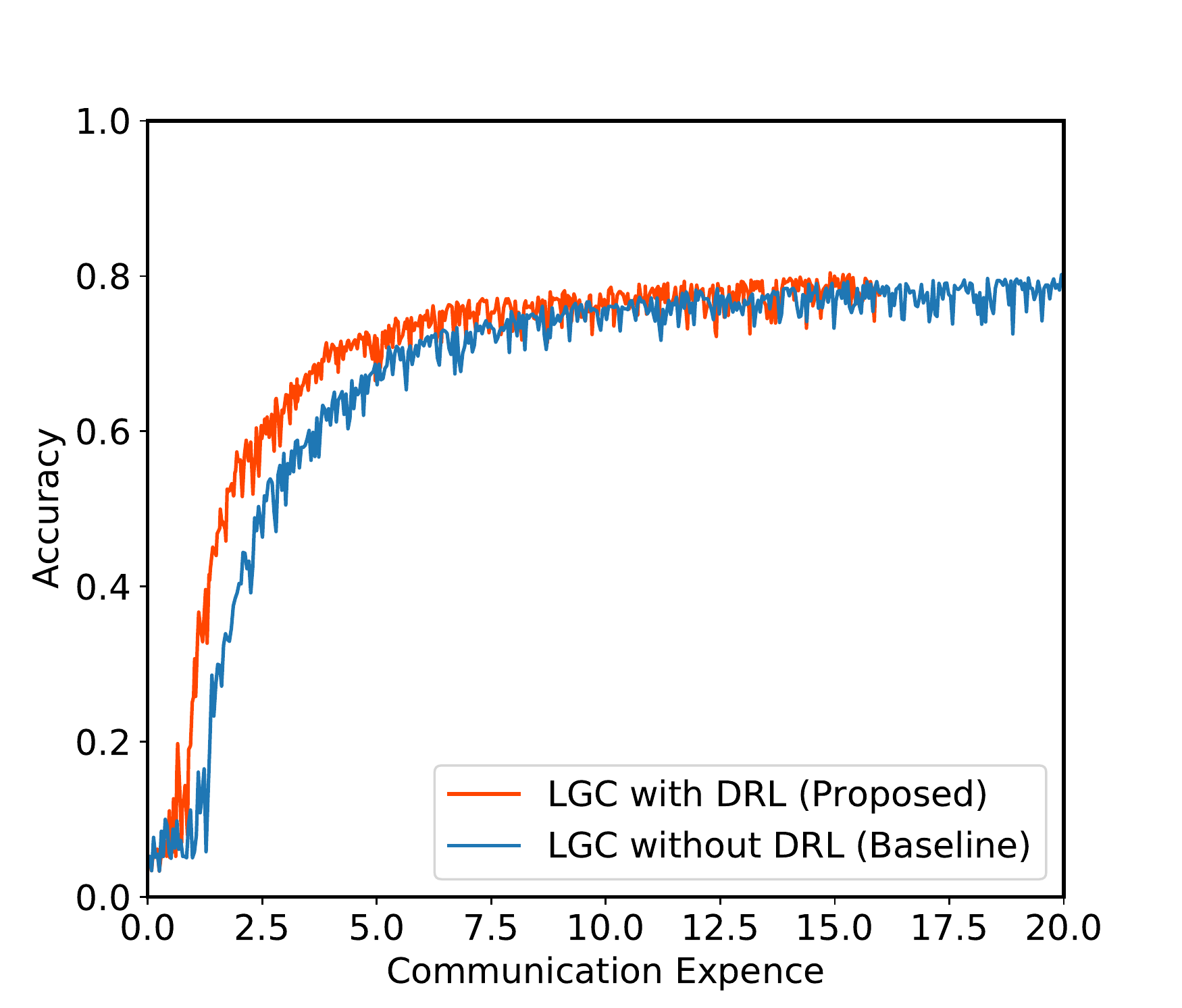}
        \end{minipage}
    }
\caption{Convergence curves of different mechanisms with CNN on MNIST.}
\label{fig:cnn_on_mnist}
\end{figure*}

\para{{Hyperparameters Setting. }}
\label{subsubsec_hyperparameters_setting}
For all experiments, we set the learning rate and batch size as $0.01$ and $64$. By default, we employ 3 devices and consider 3 different communication channels for FL. 
To quantify the energy cost for different channels, we adopt a Gaussian distribution with mean and standard deviation values \cite{wang2019adaptive}.
The parameters of this distribution are given in Table \ref{tb:resource_consumption}.

\begin{table}[!htbp]
    \caption{Energy consumption for different communcation channels.}
    \centering
    \begin{tabular}{|c|c|c|}
        \hline
        Channel Type & Mean (J/MB) & Standard Deviation \\ 
        \hline
        3G & $1296$ & 0.00033 \\ 
        4G & $2.2 \times 1296$ & 0.00033 \\
        5G & $2.5 \times 2.2 \times 1296$ & 0.00033 \\
        \hline
    \end{tabular}
    \label{tb:resource_consumption}
\end{table}

\subsection{Experiment Results}
\label{subsec:Simuresults}

\begin{figure}[h]
    \centering
    \subfigure[Loss vs. episode]{
        \label{fig:drl_loss}
        \begin{minipage}[t]{0.46\linewidth}
            \centering
            \includegraphics[width=1.85in]{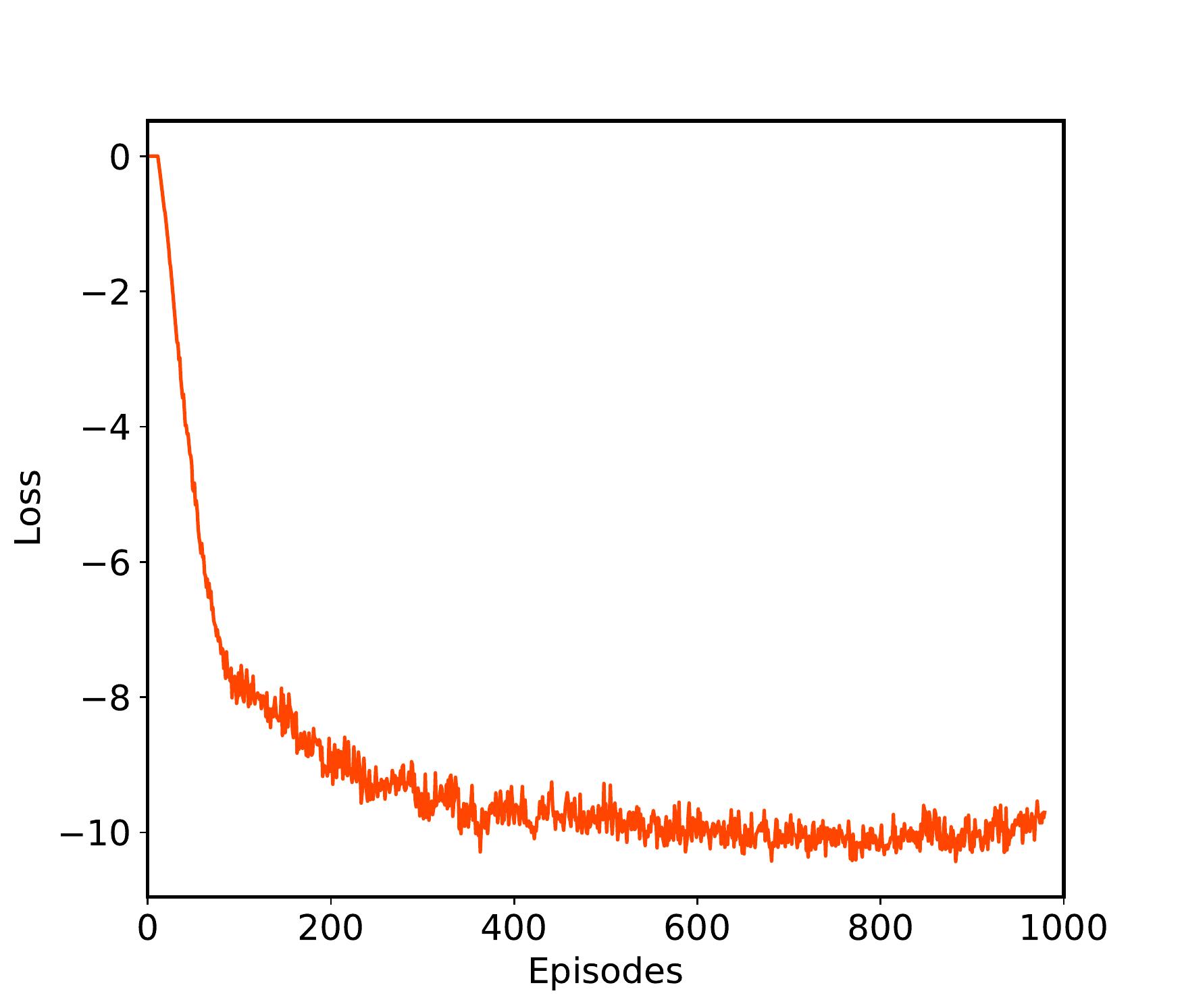}
        \end{minipage}
    }
    \subfigure[Reward vs. episode]{
        \label{fig:drl_reward}
        \begin{minipage}[t]{0.46\linewidth}
            \centering
            \includegraphics[width=1.85in]{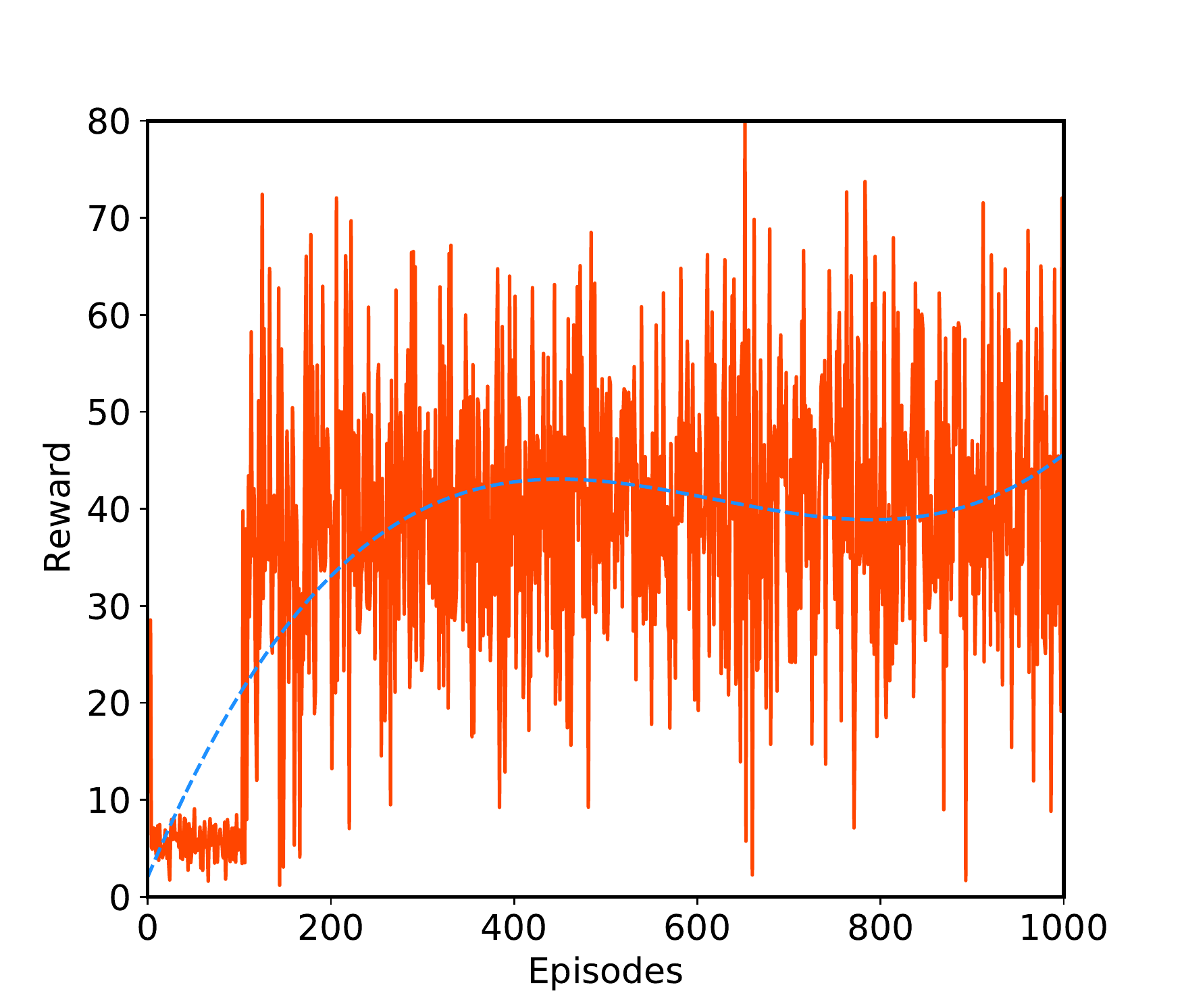}
        \end{minipage}
    }
    \caption{Convergence curves of DRL training.}
    \label{fig:drl_training}
\end{figure}

\para{Results of DRL Training. }
\label{subsubsec:SimuresultsDDPG}
The DRL training is conducted simultaneously with the FL procedure. 
In Figure~\ref{fig:drl_loss}, we first observe the change of loss with the increasing episode in DRL. 
The loss decreases quickly in the earlier stages of DRL training, because the DRL agent has no information about network condition and the FL training leads to a large training loss. Thanks to the efficient exploration and the experience replay, the reward will rapidly decrease with the model training procession. 
Figure~\ref{fig:drl_reward} shows the change of reward.
Specifically, the reward value increases with the epochs, because the DRL model can learn a better policy to achieve a better reward.

\begin{figure*}[htbp]
\centering
    \subfigure[Loss vs. epochs]{
        \label{fig:reward}
        \begin{minipage}[t]{0.23\linewidth}
            \centering
            \includegraphics[width=1.8in]{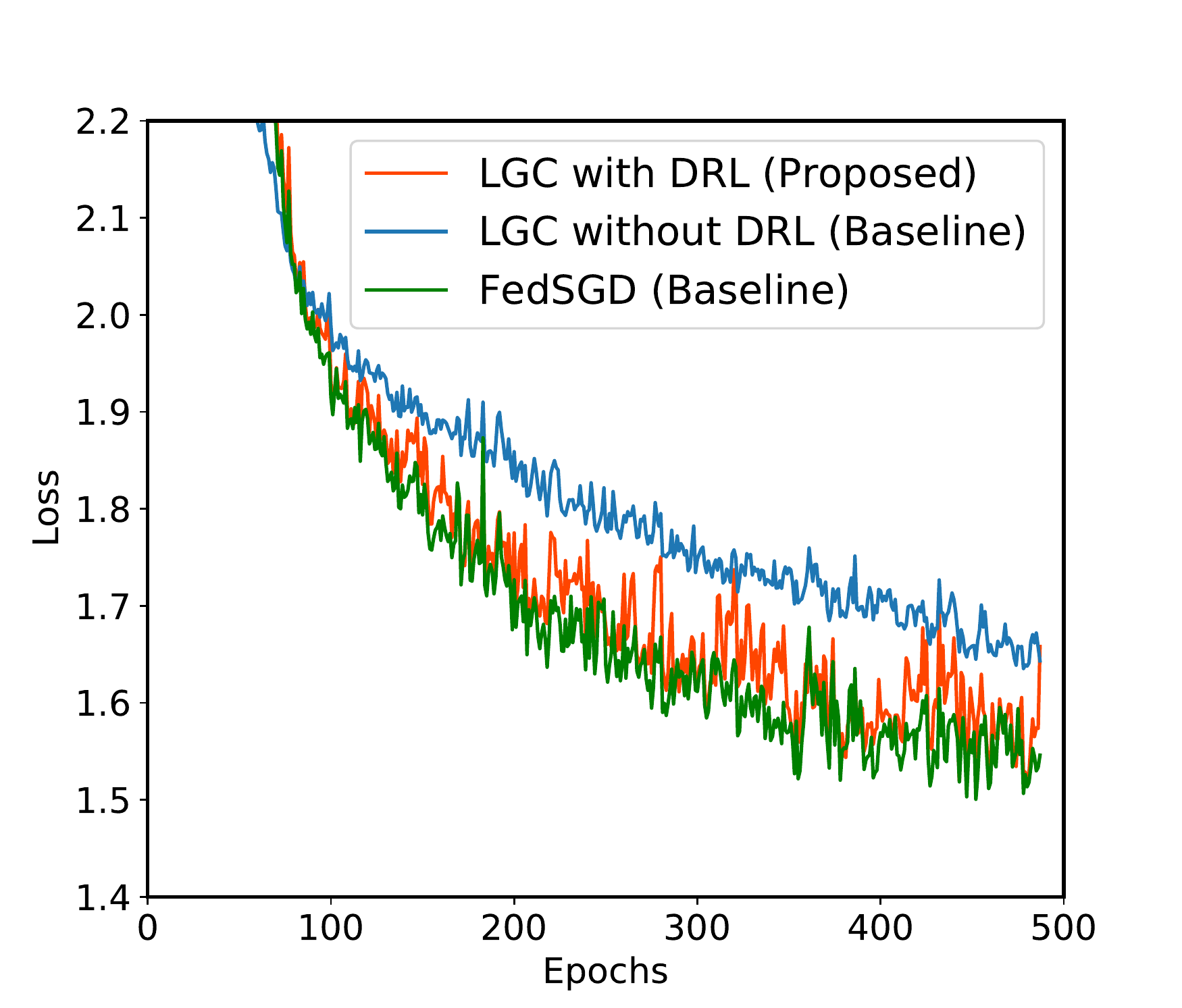}
        \end{minipage}
    }
    \subfigure[Accuracy vs. epochs]{
        \label{fig:mnist_lr}
        \begin{minipage}[t]{0.23\linewidth}
            \centering
            \includegraphics[width=1.8in]{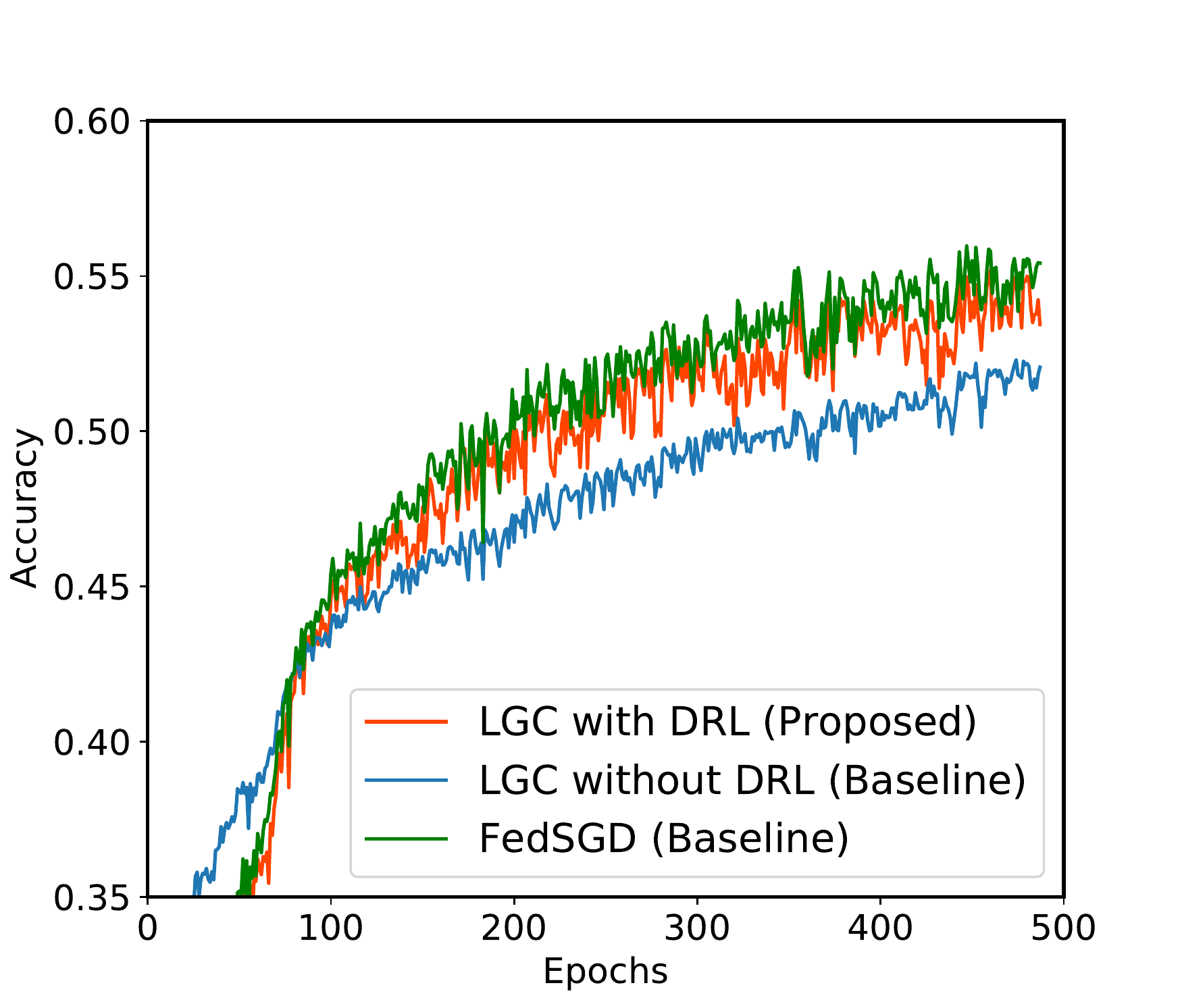}
        \end{minipage}
    }
    \subfigure[Accuracy vs. energy consumption]{
        \label{fig:cifar10_lenet}
        \begin{minipage}[t]{0.23\linewidth}
            \centering
            \includegraphics[width=1.8in]{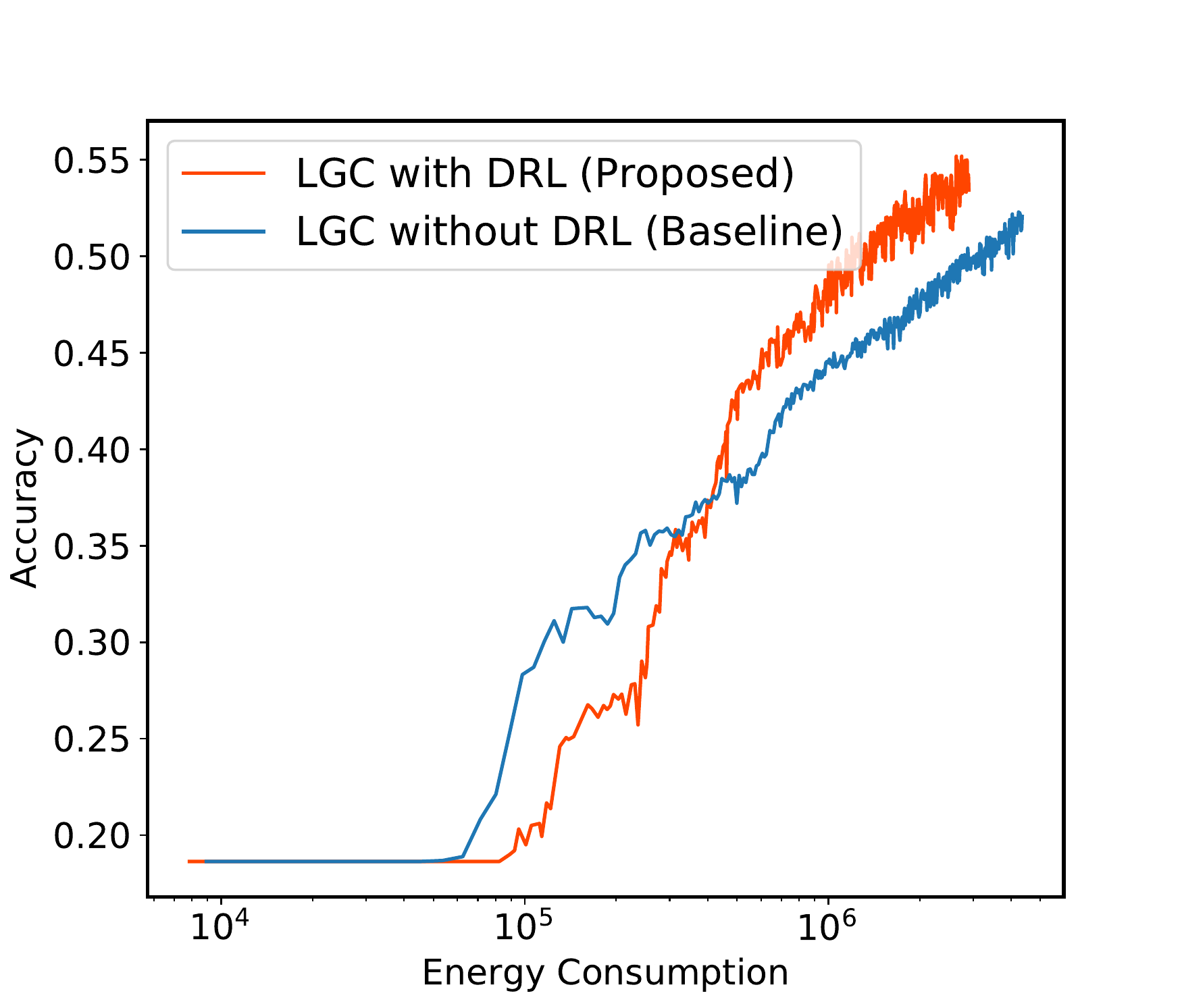}
        \end{minipage}
    }     
    \subfigure[Accuracy vs. money cost]{
        \label{fig:mnist_lenet}
        \begin{minipage}[t]{0.23\linewidth}
            \centering
            \includegraphics[width=1.8in]{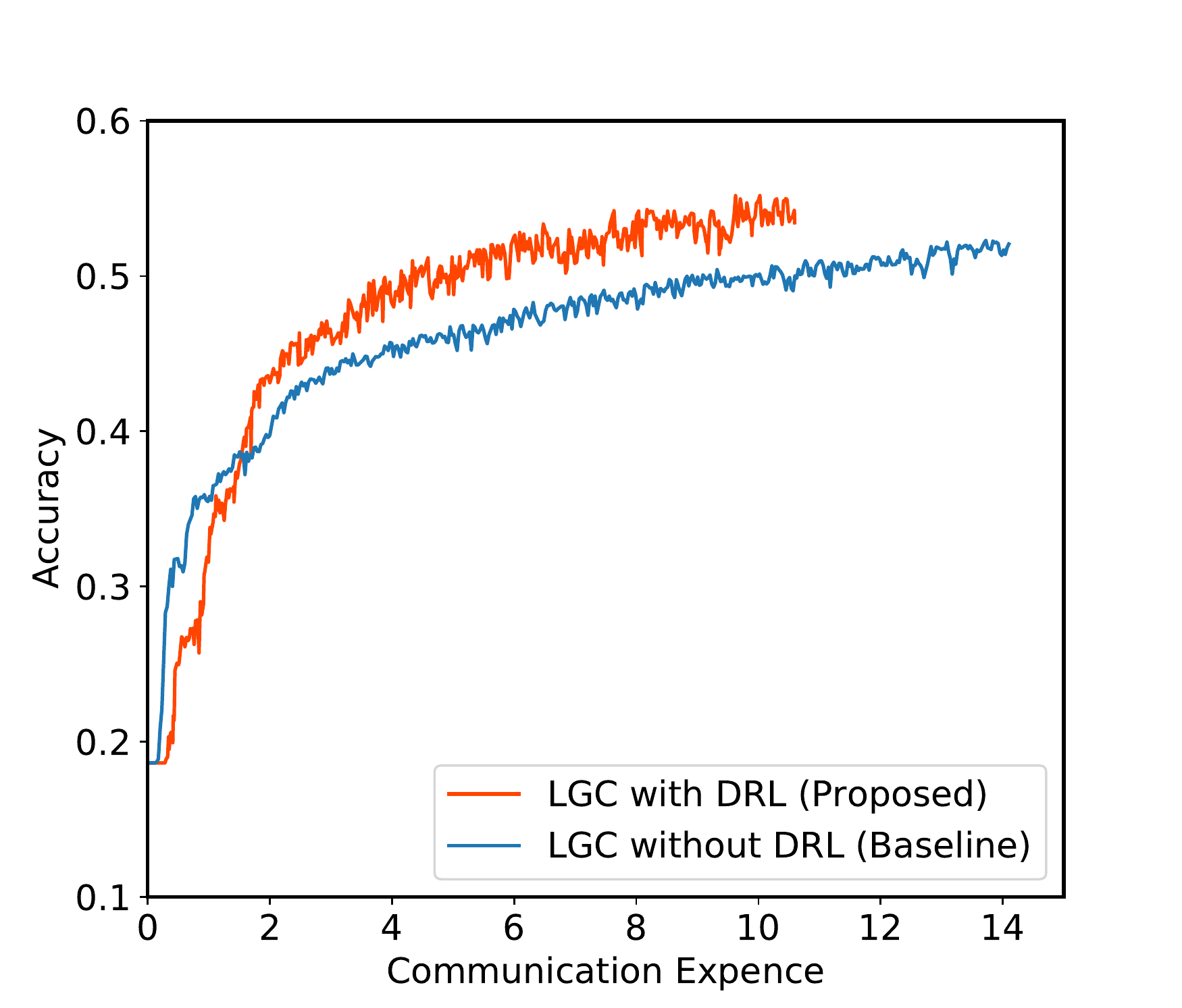}
        \end{minipage}
    }    
\caption{Convergence curves of different mechanisms with RNN on Shakespeare.}
\label{fig:rnn_on_shakespeare}
\end{figure*}

\para{Results of Performance.} We compare \system{} to baselines with different datasets and models. The convergence curves of loss and model accuracy are shown in the first two plots of Figure~\ref{fig:lr_on_mnist},
Figure~\ref{fig:cnn_on_mnist} and Figure~\ref{fig:rnn_on_shakespeare}.
We can find that \system{} convergences with a similar rate with the baselines and \system{} has very little impact on the best model accuracy.
We also compare \system{} to baselines with energy and money budgets.
By the results from the last two plots of Figure~\ref{fig:lr_on_mnist},
Figure~\ref{fig:cnn_on_mnist} and Figure~\ref{fig:rnn_on_shakespeare}, \system{} can greatly reduce the energy and money when achieving the target accuracy.
The reason for the significant performance improvement of \system{} under the resource budgets is that \system{} performs communication compression and employs multi-channel communication between edge nodes and the edge server, and the DRL based control algorithm can dynamically adjust its local computation and communication decisions.

\section{Related Works}
\label{sec:related}








These unique characteristics of FL lead to mainly two practical issues in FL implementation, i.e., (i) communication cost (ii) resource allocation. 
In this section, we review related works that address each of these issues.

\subsection{Communication Cost}

\para{Local Computation.}
Some recent works propose to perform more computation on edge nodes before each global aggregation to reduce the number of communication rounds needed for the model training \cite{mcmahan2017communication,yao2018two,liu2020client}. 
However, these approaches may increase computation cost and delay convergence if global aggregation is too infrequent. The tradeoff between these sacrifices and communication cost reduction thus has to be well-managed.

\para{Gradient Compression.}
To reduce the traffic per communication round instead of the number of communication rounds, some other works let each participant communicate the compressed gradients rather than raw gradients for every global synchronization by quantization \cite{wen2017terngrad,alistarh2017qsgd} or sparsification \cite{wangni2017gradient,stich2018sparsified,basu2019qsparse}.
However, these studies often ignored the heterogeneity among mobile devices (e.g., in computing capabilities and communication bandwidth) and required identical compression levels across all the participants and thereby exhibiting
less flexibility.

\subsection{Resource Allocation}

\para{Adaptive Aggregation.} 
In recent works, adaptive adjustment of global aggregation frequency has been investigated to increase training efficiency subject to resource constraints \cite{sprague2018asynchronous,wang2019adaptive}.
\cite{sprague2018asynchronous} proposed asynchronous FL where model aggregation occurs whenever local updates are received by the FL server.
\cite{wang2019adaptive} proposed to use adaptive global aggregation frequency based on resource constraints.
While properly managing the system resources to enable FL in mobile edge networks, these studies overlook reducing resource consumption intrinsically in the essence of learning algorithm itself, thus hindering the substantial boost in training efficiency and resource utilization.

\para{Joint Communication Techniques and Resource Management.}
Even though computation capabilities of mobile devices have grown rapidly, many devices still face a scarcity of radio resources \cite{jordan2018communication}. Given that local model transmission is an integral part of FL, there has been a growing number of studies that focus on developing novel wireless communication techniques for efficient FL \cite{amiri2020federated,yang2020federated}.
However, signal distortion can lead to a drop in accuracy, and the scalability is also an issue when large heterogeneous networks are involved.
On a higher level, wireless technologies, such as IEEE 802.11a and 5G, provide multiple non-overlapping channels.
The available network capacity can be increased by using multiple channels, and nodes can be equipped with multiple interfaces to utilize the available channels. 
\section{Conclusion}
\label{sec:conclusion}
We tackle the diverse resource utilization issue of FL by proposing LGC, a  multi-channel transmission and communication compression co-designed framework in a severely limited  MEC scenario. We analyze a convergence upper bound on LGC' results and design a learning-based control algorithm for per-device to dynamically decide dynamically its local computation and communication decisions in each epoch. The experimental results demonstrate that the LGC framework can perform better than baselines.


\bibliography{aaai22}

 \section{Appendix}
\label{appendix}

Inspired by the perturbed iterate analysis framework \cite{mania2015perturbed}, we define virtual sequences for every device $m \in \mathcal{M}$ and for all $t \geq 0$ as follows:

\begin{equation}
\label{equation_virtual_sequences}
\begin{aligned}
\widetilde{\mathbf{w}}^{(0)}_{m} 
:= \widehat{\mathbf{w}}^{(0)}_{m}
\ \ \
\widetilde{\mathbf{w}}^{(t+1)}_{m} 
:= \widetilde{\mathbf{w}}^{(t)}_{m} - \eta^{(t)} \nabla f_{m}(\widehat{\mathbf{w}}^{(t)}_{m}; \mathcal{D}^{(t)}_{m})
\end{aligned}
\end{equation}

We also define

\begin{equation}
\begin{aligned}
\mathbf{q}^{(t)}
& := \frac{1}{M} \sum_{m=1}^M \nabla f_{m}(\widehat{\mathbf{w}}^{(t)}_{m}; \mathcal{D}^{(t)}_{m}) \\
\overline{\mathbf{q}}^{(t)}
& := \mathbb{E}_{\mathcal{D}_{\mathcal{M}}^{(t)}}[\mathbf{q}^{(t)}] = \frac{1}{M} \sum_{m=1}^M \nabla f_{m}(\widehat{\mathbf{w}}^{(t)}_{m}) \\
\widetilde{\mathbf{w}}^{(t+1)}
& := \frac{1}{M} \sum_{m=1}^{M} \widetilde{{\mathbf{w}}}^{(t+1)}_{m}
= \widetilde{\mathbf{w}}^{(t)} - \eta^{(t)} \mathbf{q}^{(t)} \\
\widehat{\mathbf{w}}^{(t)}
& := \frac{1}{M} \sum_{m=1}^{M} \widehat{\mathbf{w}}^{(t)}_{m} \\
\mathcal{I}_{m} 
& = \left\{t^{(i)}_{m}: i \in \mathbb{Z}^{+}, t^{(i)}_{m} \in\mathcal{T},\left|t^{(i)}_{m}-t^{(j)}_{m}\right| \leq H, \forall|i-j| \leq 1\right\} \\
\end{aligned}
\end{equation}

\subsection{Proof of Theorem \ref{theorem_async_convergence_rate}}

\textit{Proof.}
Let $\mathbf{w}^{*}$ be the minimizer of $f(\mathbf{w})$, therefore we have $\nabla f\left(\mathbf{w}^{*}\right)=0$. We denote $f\left(\mathbf{w}^{*}\right)$ by $f^{*}$. By taking the average of the virtual sequences $\widetilde{\mathbf{w}}^{(t+1)}_{m}=\widetilde{\mathbf{w}}_{m}^{(t)}-\eta^{(t)} \nabla f_{m}(\widehat{\mathbf{w}}^{(t)}_{m}; \mathcal{D}^{(t)}_{m})$ for each worker
$m \in\mathcal{M}$ and defining $\mathbf{q}^{(t)}
 := \frac{1}{M} \sum_{m=1}^M \nabla f_{m}(\widehat{\mathbf{w}}^{(t)}_{m}; \mathcal{D}^{(t)}_{m})$, we get

\begin{equation}
\label{w_tilde_tplus1}
\widetilde{\mathbf{w}}^{(t+1)}=\widetilde{\mathbf{w}}^{(t)}-\eta^{(t)} \mathbf{q}^{(t)}
\end{equation}

Define $\mathcal{D}_{\mathcal{M}}^{(t)}$ as the set of random sampling of the mini-batches at each worker $\left\{\mathcal{D}_{1}^{(t)}, \mathcal{D}_{2}^{(t)}, \ldots, \mathcal{D}_{M}^{(t)}\right\}$ and let $\overline{\mathbf{q}}^{(t)}
 = \mathbb{E}_{\mathcal{D}_{\mathcal{M}}^{(t)}}[\mathbf{q}^{(t)}] = \frac{1}{M} \sum_{m=1}^M \nabla f_{m}(\widehat{\mathbf{w}}^{(t)}_{m})$. From (\ref{w_tilde_tplus1}) we can get

\begin{equation}
\begin{aligned}
\label{w_tilde_tplus1_minus_w_star}
\left\|\widetilde{\mathbf{w}}^{(t+1)}-\mathbf{w}^{*}\right\|^{2}
& =\left\|\widetilde{\mathbf{w}}^{(t)}-\eta^{(t)}{\mathbf{q}}^{(t)}-\mathbf{w}^{*}-\eta^{(t)} \overline{\mathbf{q}}^{(t)}+\eta^{(t)} \overline{\mathbf{q}}^{(t)}\right\|^{2}
\\&=\left\|\widetilde{\mathbf{w}}^{(t)}-\mathbf{w}^{*}-\eta^{(t)} \overline{\mathbf{q}}^{(t)}\right\|^{2}
\\&+(\eta^{(t)})^{2}\left\|\mathbf{q}^{(t)}-\overline{\mathbf{q}}^{(t)}\right\|^{2}
\\&-2 \eta^{(t)}\left\langle\widetilde{\mathbf{w}}^{(t)}-\mathbf{w}^{*}-\eta^{(t)} \overline{\mathbf{q}}^{(t)}, \mathbf{q}^{(t)}-\overline{\mathbf{q}}^{(t)}\right\rangle
\end{aligned}    
\end{equation}

Taking the expectation w.r.t. the sampling $\mathcal{D}_\mathcal{M}^{(t)}$ at time $t$ (conditioning on the past) and noting that last term in (\ref{w_tilde_tplus1_minus_w_star}) becomes zero gives:

\begin{equation}
\begin{aligned}
\mathbb{E}_{\mathcal{D}_{\mathcal{M}}^{(t)}}\left\|\widetilde{\mathbf{w}}^{(t+1)}-\mathbf{w}^{*}\right\|^{2}
&=\left\|\widetilde{\mathbf{w}}^{(t)}-\mathbf{w}^{*}-\eta^{(t)} \overline{\mathbf{q}}^{(t)}\right\|^{2}
\\&+(\eta^{(t)})^{2}\left\|\mathbf{q}^{(t)}-\overline{\mathbf{q}}^{(t)}\right\|^{2}
\end{aligned}
\end{equation}

It follows from the Jensen's inequality and independence that $\mathbb{E}_{\mathcal{D}_{\mathcal{M}}^{(t)}}\|\mathbf{q}^{(t)}-\overline{\mathbf{q}}^{(t)}\|^{2} \leq \frac{\sum_{m=1}^{M} \sigma_{m}^{2}}{b M^{2}}$. This gives

\begin{equation}
\begin{aligned}
\mathbb{E}_{\mathcal{D}_{\mathcal{M}}^{(t)}}\left\|\widetilde{\mathbf{w}}^{(t+1)}-\mathbf{w}^{*}\right\|^{2}
&=\left\|\widetilde{\mathbf{w}}^{(t)}-\mathbf{w}^{*}-\eta^{(t)} \overline{\mathbf{q}}^{(t)}\right\|^{2}
\\&+(\eta^{(t)})^{2}
\frac{\sum_{m=1}^{M} \sigma_{m}^{2}}{b M^2}
\end{aligned}
\end{equation}

Now we bound the first term on the RHS. Using $\mu$-strong convexity and $L$-smoothness of $f$, together with some algebraic manipulations provided in Lemma 14 in \cite{basu2019qsparse}, we arrive at

\begin{equation}
\begin{aligned}
\label{equation_bound_v1_2}
\mathbb{E}\|\widetilde{\mathbf{w}}^{(t+1)}-\mathbf{w}^{*}\|_{2}^{2} 
&\leq(1-\frac{\mu \eta^{(t)}}{2}) \mathbb{E}\|\widetilde{\mathbf{w}}^{(t)}-\mathbf{w}^{*}\|_{2}^{2}
\\&-\frac{\eta^{(t)} \mu}{2 L} (\mathbb{E}\left[f\left(\widehat{\mathbf{w}}^{(t)}\right)\right]-f^{*})
\\&+\eta^{(t)}(\frac{3 \mu}{2}+3 L) \mathbb{E}\|\widehat{\mathbf{w}}^{(t)}-\widetilde{\mathbf{w}}^{(t)}\|_{2}^{2} \\
\\&+\frac{3 \eta^{(t)} L}{M} \sum_{m=1}^{M} \mathbb{E}\|\widehat{\mathbf{w}}^{(t)}-\widehat{\mathbf{w}}^{(t)}_{m}\|_{2}^{2}
\\&+(\eta^{(t)})^{2} \frac{\sum_{m=1}^{M} \sigma_{m}^{2}}{b M^{2}}
\end{aligned}
\end{equation}

Now we have to bound the deviation of local sequences $\frac{1}{M} \sum_{m=1}^{M} \mathbb{E}\left\|\widehat{\mathbf{w}}^{(t)}-\widehat{\mathbf{w}}_{m}^{(t)}\right\|_{2}^{2}$ and the difference between the virtual and true sequences $\mathbb{E}\left\|\widehat{\mathbf{w}}^{(t)}-\widetilde{\mathbf{w}}^{(t)}\right\|_{2}^{2}$.
We show these below in Lemma \ref{lemma_contracting_local_sequence_deviation} and Lemma \ref{lemma_contracting_distance_between_virtual_and_true_sequence}.

\begin{lemma}
\label{lemma_contracting_local_sequence_deviation}
(Contracting local sequence deviation). Let $\operatorname{gap}\left(\mathcal{I}_{m}\right) \leq H$ holds for every $m \in \mathcal{M}$. For $\widehat{\mathbf{w}}_{t}^{(r)}$ generated according to Algorithm \ref{sync-ef-compressed-local-sgd} with decaying learning rate $\eta^{(t)}$ and letting $\widehat{\mathbf{w}}^{(t)}=$ $\frac{1}{M} \sum_{m=1}^{M} \widehat{\mathbf{w}}^{(t)}_{m}$, we have the following bound on the deviation of the local sequences:

\begin{equation}
\label{equation_contracting_local}
\frac{1}{M} \sum_{m=1}^{M} \mathbb{E}\left\|\widehat{\mathbf{w}}^{(t)}-\widehat{\mathbf{w}}_{m}^{(t)}\right\|_{2}^{2} \leq
8\left(1+C^{\prime \prime} H^{2}\right) (\eta^{(t)})^{2} G^{2} H^{2}    
\end{equation}

where $C^{\prime \prime}=\frac{8}{M}\sum_{m=1}^{M}(4-2 \gamma_m)\left(1+\frac{C}{\gamma_m^{2}}\right)$ and $C$ is a constant satisfying $C \geq \min_{m\in\mathcal{M}} \frac{4 a \gamma_m\left(1-\gamma_m^{2}\right)}{a \gamma_m-4 H}$.
\end{lemma}

\begin{lemma}
\label{lemma_contracting_distance_between_virtual_and_true_sequence}
(Contracting distance between virtual and true sequence). Let $\operatorname{gap}\left(\mathcal{I}_{m}\right) \leq H$ holds for every $m \in \mathcal{M}$. If we run Algorithm \ref{sync-ef-compressed-local-sgd} with a decaying learning rate $\eta^{(t)}$, then we have the following bound on the difference between the true and virtual sequences:

\begin{equation}
\label{equation_contracting_distance_between_virtual_and_true_sequence}
\mathbb{E}\left\|\widehat{\mathbf{w}}^{(t)}-\widetilde{\mathbf{w}}^{(t)}\right\|_{2}^{2} \leq C^{\prime} (\eta^{(t)})^{2} H^{4} G^{2}+12 C \frac{(\eta^{(t)})^{2}}{\gamma^{2}} G^{2} H^{2} 
\end{equation}

where $C^{\prime}=192C^{\prime \prime}=\frac{8}{M}\sum_{m=1}^{M}(4-2 \gamma_m)\left(1+\frac{C}{\gamma_m^{2}}\right)$ and $C$ is a constant satisfying $C \geq \min_{m\in\mathcal{M}} \frac{4 a \gamma_m\left(1-\gamma_m^{2}\right)}{a \gamma_m-4 H}$.
\end{lemma}

Substituting the bounds from (\ref{equation_contracting_local})(\ref{equation_contracting_distance_between_virtual_and_true_sequence}) into (\ref{equation_bound_v1_2}) yields

\begin{equation}
\begin{aligned}
&\mathbb{E}\|\widetilde{\mathbf{w}}^{(t+1)}-\mathbf{w}^{*}\|_{2}^{2} \\
&\leq(1-\frac{\mu \eta^{(t)}}{2}) \mathbb{E}\|\widetilde{\mathbf{w}}^{(t)}-\mathbf{w}^{*}\|_{2}^{2}
\\&-\frac{\eta^{(t)} \mu}{2 L} \varepsilon^{(t)}
\\&+\eta^{(t)}(\frac{3 \mu}{2}+3 L) [C^{\prime} (\eta^{(t)})^{2} H^{4} G^{2}+12 C \frac{(\eta^{(t)})^{2}}{\gamma^{2}} G^{2} H^{2}] \\
\\&+\frac{\left(3 \eta^{(t)} L\right)}{M} [8\left(1+C^{\prime \prime} H^{2}\right) (\eta^{(t)})^{2} G^{2} H^{2}]
\\&+(\eta^{(t)})^{2} \frac{\sum_{m=1}^{M} \sigma_{m}^{2}}{b M^{2}}
\end{aligned}
\end{equation}

Employing a slightly modified result than Lemma 3.3 in \cite{stich2018sparsified} with $a^{(t)}=\mathbb{E}\left\|\widetilde{\mathbf{w}}^{(t)}-\mathbf{w}^{*}\right\|_{2}^{2}, A=$ $\frac{\sum_{m=1}^{M} \sigma_{m}^{2}}{b M^{2}}$ and $B=(\frac{3\mu}{2} + 3L) (\frac{12CG^2 H^2}{\gamma^2} + C_{1} (\eta^{(t)})^2 H^4G^2) + 24(1+C_{2}H^2) LG^2H^2$, we have
$$
a^{(t+1)} \leq\left(1-\frac{\mu \eta^{(t)}}{2}\right) a^{(t)}-\frac{\mu \eta^{(t)}}{2 L} \varepsilon^{(t)}+(\eta^{(t)})^{2} A+(\eta^{(t)})^{3} B .
$$
For $\eta^{(t)}=\frac{8}{\mu(a+t)}$ and $s^{(t)}=(a+t)^{2}, S=\sum_{t=0}^{T-1} s^{(t)} \geq \frac{T^{3}}{3}$, we have
$$
\frac{\mu}{2 L S} \sum_{t=0}^{T-1} s^{(t)} \varepsilon^{(t)} \leq \frac{\mu a^{3}}{8 S} a^{(0)}+\frac{4 T(T+2 a)}{\mu S} A+\frac{64 T}{\mu^{2} S} B
$$
From convexity, we can finally write
$$
\mathbb{E} f\left(\overline{\mathbf{w}}^{(T)}\right)-f^{*} \leq \frac{L a^{3}}{4 S} a^{(0)}+\frac{8 L T(T+2 a)}{\mu^{2} S} A+\frac{128 L T}{\mu^{3} S} B
$$
Where $\overline{\mathbf{w}}^{(T)}:=\frac{1}{S} \sum_{t=0}^{T-1}\left[s^{(t)}\left(\frac{1}{M} \sum_{m=1}^{M} \widehat{\mathbf{w}}^{(t)}_{m}\right)\right]=\frac{1}{S} \sum_{t=0}^{T-1} s^{(t)} \widehat{\mathbf{w}}^{(t)}$. This completes the proof of Theorem \ref{theorem_async_convergence_rate}.

\subsection{Proof of Lemma \ref{lemma_contracting_local_sequence_deviation}}




\textit{Proof.}
Fix a time $t$ and consider any worker $m \in\mathcal{M}$. Let $t_{m} \in \mathcal{I}_{m}$ denote the last synchronization step until time $t$ for the $m$'th worker. Define $t_{0}^{\prime}:=\min _{m \in\mathcal{M}} t_{m} .$ We need to upper-bound $\frac{1}{M} \sum_{m=1}^{M} \mathbb{E}\left\|\widehat{\mathbf{w}}^{(t)}-\widehat{\mathbf{w}}^{(t)}_{m}\right\|^{2}$. Note that for any $M$ vectors $\mathbf{u}_{1}, \ldots, \mathbf{u}_{M}$, if we let $\overline{\mathbf{u}}=\frac{1}{M} \sum_{i=1}^{M} \mathbf{u}_{i}$, then
$\sum_{i=1}^{n}\left\|\mathbf{u}_{i}-\overline{\mathbf{u}}\right\|^{2} \leq \sum_{i=1}^{M}\left\|\mathbf{u}_{i}\right\|^{2}$. We use this in the first inequality below.

\begin{equation}
\label{equation_2_bounds}
\begin{aligned}
&\frac{1}{M} \sum_{m=1}^{M} \mathbb{E}\left\|\widehat{\mathbf{w}}^{(t)}-\widehat{\mathbf{w}}^{(t)}_{m}\right\|^{2}\\ &=\frac{1}{M} \sum_{m=1}^{M} \mathbb{E}\left\|\widehat{\mathbf{w}}^{(t)}_{m}-\overline{\bar{\mathbf{w}}}^{(t_{0}^{\prime})}-\left(\widehat{\mathbf{w}}^{(t)}-\overline{\bar{\mathbf{w}}}^{(t_{0}^{\prime})}\right)\right\|^{2} \\
& \leq \frac{1}{M} \sum_{m=1}^{M} \mathbb{E}\left\|\widehat{\mathbf{w}}_{m}^{(t)}-\overline{\bar{\mathbf{w}}}^{(t_{0}^{\prime})}\right\|^{2} \\
& \leq \frac{2}{M} \sum_{m=1}^{M} \mathbb{E}\left\|\widehat{\mathbf{w}}_{m}^{(t)}-\widehat{\mathbf{w}}^{(t_{m})}_{m}\right\|^{2}+\frac{2}{M} \sum_{m=1}^{M} \mathbb{E}\left\|\widehat{\mathbf{w}}^{(t_{m})}_{m}-\overline{\bar{\mathbf{w}}}^{(t_{0}^{\prime})}\right\|^{2}
\end{aligned}    
\end{equation}

We bound both the terms separately. For the first term:

\begin{equation}
\label{equation_2_bounds_1}
\begin{aligned}
\mathbb{E}\left\|\widehat{\mathbf{w}}^{(t)}_{m}-\widehat{\mathbf{w}}^{(t_{m})}_{m}\right\|^{2} &=\mathbb{E}\left\|\sum_{j=t_{m}}^{t-1} \eta^{(j)} \nabla f_{\mathcal{D}^{(j)}_{m}}\left(\widehat{\mathbf{w}}^{(j)}_{m}\right)\right\|^{2} \\
& \leq\left(t-t_{m}\right) \sum_{j=t_{m}}^{t-1} \mathbb{E}\left\|\eta^{(j)} \nabla f_{\mathcal{D}^{(j)}_{m}}\left(\widehat{\mathbf{w}}^{(j)}_{m}\right)\right\|^{2} \\
& \leq\left(t-t_{m}\right)^2 {\eta^{t_{m}} G}^2 \\
& \leq 4 (\eta^{(t)})^{2} H^{2} G^{2}
\end{aligned}
\end{equation}

The last inequality (\ref{equation_2_bounds_1}) uses $\eta^{(t_{m})} \leq 2 \eta^{(t_{m}+H)} \leq 2 \eta^{(t)}$ and $t-t_{m} \leq H$. To bound the second term of (\ref{equation_3_bounds}), note that we have

\begin{equation}
\label{equation_2_bounds_2}
\bar{\bar{\mathbf{w}}}^{t_{m}}_{m}=\bar{\bar{\mathbf{w}}}^{(t_{0}^{\prime})}-\frac{1}{M} \sum_{s=1}^{M} \sum_{j=t_{0}^{\prime}}^{t_{m}-1} \mathbbm{1}\left\{j+1 \in \mathcal{I}_{s}\right\} g_{s}^{(j)}
\end{equation}

Note that $\widehat{\mathbf{w}}^{(t_{m})}_{m}=\overline{\overline{\mathbf{w}}}^{(t_{m})}_{m}$, because at synchronization steps, the local parameter vector becomes equal to the global parameter vector. Using this, the Jensen's inequality, and that $\| \mathbbm{1}\{j+1 \in$ $\left.\mathcal{I}_{s}\right\} g^{(j)}_{s}\left\|^{2} \leq\right\| g^{(j)}_{s} \|^{2}$, we can upper-bound (\ref{equation_2_bounds_2}) as

\begin{equation}
\label{equantion_bound_hattr_oot0prime}
\mathbb{E}\left\|\widehat{\mathbf{w}}^{t_{m}}_{m}-\overline{\overline{\mathbf{w}}}^{(t_{0}^{\prime})}\right\|^{2} \leq \frac{\left(t_{m}-t_{0}^{\prime}\right)}{M} \sum_{s=1}^{M} \sum_{j=t_{0}^{\prime}}^{t_{m}} \mathbb{E}\left\|g^{(j)}_{s}\right\|^{2}  
\end{equation}

Now we bound $\mathbb{E}\left\|g^{(j)}_{s}\right\|^{2}$ for any $j \in\left\{t_{0}^{\prime}, \ldots, t_{m}\right\}$ and $s \in\mathcal{M}$: Since $\mathbb{E}\left\|\mathcal{C}_{m}^{(t)}(\mathbf{u})\right\|^{2} \leq B\|\mathbf{u}\|^{2}$
holds for every $\mathbf{u}$, with $B=(4-2 \gamma_{m})$ \footnote{This can be seen as follows: $
\mathbb{E}\|\mathcal{C}_{m}^{(t)}(\mathbf{u})\|^2 \leq 2 \mathbb{E}\|\mathbf{u} - \mathcal{C}_{m}^{(t)}(\mathbf{u})|^2 + 2\|\mathbf{u}\|^2 
\leq 2(1-\gamma_{m})\|\mathbf{u}\|^2 + 2\|\mathbf{u}\|^2
$}, we have for any $s \in\mathcal{M}$ that

\begin{equation}
\label{equation_bound_g}
\begin{aligned}
\mathbb{E}\left\|\mathbf{g}^{(j)}_{s}\right\|^{2} & \leq B \mathbb{E}\left\|\mathbf{e}^{(j)}_{s}+\mathbf{w}_{s}^{(j)}-\widehat{\mathbf{w}}^{(j+\frac{1}{2})}_{s}\right\|^{2} \\
& \leq 2 B \mathbb{E}\left\|\mathbf{e}^{(j)}_{s}\right\|^{2}+2 B \mathbb{E}\left\|\mathbf{w}^{(j)}_{s}-\widehat{\mathbf{w}}^{(j+\frac{1}{2})}_{s}\right\|^{2}
\end{aligned}
\end{equation}

We can directly use Lemma \ref{lemma_memory_contraction} to bound the first term in (\ref{equation_bound_g}) as $\mathbb{E}\left\|\mathbf{e}_{j}^{(s)}\right\|^{2} \leq 4 C \frac{(\eta^{(j)})^{2}}{\gamma_m^{2}} H^{2} G^{2}$. 
In order to bound the second term of $(\ref{equation_bound_g})$, note that $\mathbf{w}^{(j)}_{s}=\widehat{\mathbf{w}}^{(t_{s})}_{s}$, which implies that $\left\|\mathbf{w}^{(j)}_{s}-\widehat{\mathbf{w}}^{(j+\frac{1}{2})}_{s}\right\|^{2}=\left\|\sum_{l=t_{s}}^{j} \eta^{(l)} \nabla f_{\mathcal{D}^{(l)}_{s}}\left(\widehat{\mathbf{w}}_{l}^{(s)}\right)\right\|^{2}$
Taking expectation yields $\mathbb{E}\left\|\mathbf{w}^{(j)}_{s}-\widehat{\mathbf{w}}^{(j+\frac{1}{2})}_{s}\right\|^{2} \leq 4 (\eta^{(t_{s})})^{2} H^{2} G^{2} \leq 4 (\eta^{(t_{0}^{\prime})})^{2} H^{2} G^{2}$, where in the last inequality
we used that $t_{0}^{\prime} \leq t_{s}$. Using these in $(\ref{equation_bound_g})$ gives

\begin{equation}
\label{equantion_bound_g_2}
\mathbb{E}\left\|g^{(j)}_{s}\right\|^{2} \leq 8 B\left(1+\frac{C}{\gamma_m^{2}}\right) (\eta^{(t_{0}^{\prime})})^{2} H^{2} G^{2}  
\end{equation}

Since $t_{0}^{\prime} \leq t \leq t_{0}^{\prime}+H$, we have $\eta^{(t_{0}^{\prime})} \leq 2 \eta^{(t_{0}^{\prime}+H)} \leq 2 \eta^{(t)}$. Putting the bound on $\mathbb{E}\left\|\mathbf{g}^{(j)}_{s}\right\|^{2}$ (after substituting $\eta^{(t_{0}^{\prime})} \leq 2 \eta^{(t)}$ in $\left.(\ref{equantion_bound_g_2})\right)$ in $(\ref{equantion_bound_hattr_oot0prime})$ gives

\begin{equation}
\label{equation_whattrr_minus_wbarbart0prime}
\begin{aligned}
\mathbb{E}\left\|\widehat{\mathbf{w}}^{(t_{m})}_{m}-\overline{\overline{\mathbf{w}}}^{(t_{0}^{\prime})}\right\|^{2} &\leq 32 \frac{1}{M}[\sum_{m=1}^{M}(4-2\gamma_m)\left(1+\frac{C}{\gamma_m^{2}}\right)] \\
&(\eta^{(t)})^{2} H^{4} G^{2}
\end{aligned}    
\end{equation}

Putting this and the bound from $(\ref{equation_2_bounds_1})$ back in $(\ref{equation_2_bounds})$ gives

\begin{equation}
\begin{aligned}
&\frac{1}{M} \sum_{m=1}^{M} \mathbb{E}\left\|\widehat{\mathbf{w}}^{(t)}-\widehat{\mathbf{w}}^{(t)}_{m}\right\|^{2} \\
& \leq 8 (\eta^{(t)})^{2} H^{2} G^{2} \\
&+64 \frac{1}{M}[\sum_{m=1}^{M}(4-2\gamma_m)\left(1+\frac{C}{\gamma_m^{2}}\right)] (\eta^{(t)})^{2} H^{4} G^{2} \\
& \leq 8\left[1+8 [\frac{1}{M}\sum_{m=1}^{M}(4-2\gamma_m)\left(1+\frac{C}{\gamma_m^{2}}\right)] H^2 \right] (\eta^{(t)})^{2} H^{2} G^{2}
\end{aligned}
\end{equation}

This completes the proof of Lemma \ref{lemma_contracting_local_sequence_deviation}.

\subsection{Proof of Lemma \ref{lemma_contracting_distance_between_virtual_and_true_sequence}}




\textit{Proof.}
Fix a time $t$ and consider any device $m \in\mathcal{M} .$ Let $t_{m} \in \mathcal{I}_{m}$ denote the last synchronization step until time $t$ for the $m$'th device. Define $t_{0}^{\prime}:=\min _{m \in\mathcal{M}} t_{m} .$ We want to bound $\mathbb{E}\left\|\widehat{\mathbf{w}}^{(t)}-\widetilde{\mathbf{w}}^{(t)}\right\|^{2}$. 
By definition $\widehat{\mathbf{w}}^{(t)}-\widetilde{\mathbf{w}}^{(t)}=\frac{1}{M} \sum_{m=1}^{M}\left(\widehat{\mathbf{w}}_{m}^{(t)}-\widetilde{\mathbf{w}}_{m}^{(t)}\right)$. By the definition of virtual
sequences and the update rule for $\widehat{\mathbf{w}}_{m}^{(t)}$, we also have $\widehat{\mathbf{w}}^{(t)}-\widetilde{\mathbf{w}}^{(t)}=\frac{1}{M} \sum_{m=1}^{M}\left(\widehat{\mathbf{w}}^{(t_{m})}_{m}-\widetilde{\mathbf{w}}^{(t_{m})}_{m}\right)$. This can be written as

\begin{equation}
\label{whatt_minus_wtildet}
\begin{aligned}
\widehat{\mathbf{w}}^{(t)}-\widetilde{\mathbf{w}}^{(t)}
&=\left[\frac{1}{M} \sum_{m=1}^{M} \widehat{\mathbf{w}}^{(t_{m})}_{m}-\bar{\bar{\mathbf{w}}}^{(t_{0}^{\prime})}\right]
+ \left[\bar{\bar{\mathbf{w}}}^{(t_{0}^{\prime})}-\bar{\bar{\mathbf{w}}}^{(t)}\right]
\\&+\left[\bar{\bar{\mathbf{w}}}^{(t)}-\frac{1}{M} \sum_{m=1}^{M} \widetilde{\mathbf{w}}^{(t_{m})}_{m}\right] 
\end{aligned}
\end{equation}

Applying Jensen's inequality and taking expectation gives

\begin{equation}
\label{equation_3_bounds}
\begin{aligned}
\mathbb{E}\left\|\widehat{\mathbf{w}}^{(t)}-\widetilde{\mathbf{w}}^{(t)}\right\|^{2} &\leq\left[\frac{3}{M} \sum_{m=1}^{M} \mathbb{E}\left\|\widehat{\mathbf{w}}^{(t_{m})}_{m}-\bar{\bar{\mathbf{w}}}^{(t_{0}^{\prime})}\right\|^{2}\right]
\\&+\left[3 \mathbb{E}\left\|\bar{\bar{\mathbf{w}}}^{(t_{0}^{\prime})}-\bar{\bar{\mathbf{w}}}^{(t)}\right\|^{2}\right]
\\&+\left[3 \mathbb{E}\left\|\bar{\bar{\mathbf{w}}}^{(t)}-\frac{1}{M} \sum_{m=1}^{M} \widetilde{\mathbf{w}}^{(t_{m})}_{m}\right\|^{2}\right]
\end{aligned}
\end{equation}

We bound each of the three terms of (\ref{equation_3_bounds}) separately. We have upper-bounded the first term earlier in (\ref{equation_3_bounds}), which is

\begin{equation}
\mathbb{E}\left\|\widehat{\mathbf{w}}^{(t_{m})}_{m}-\bar{\bar{\mathbf{w}}}^{(t_{0}^{\prime})}\right\|^{2} \leq 32 B\left(1+\frac{C}{\gamma^{2}}\right) (\eta^{(t)})^{2} H^{4} G^{2}
\end{equation}

where $B=(4-2 \gamma)$. To bound the second term of (\ref{equation_3_bounds}), note that

\begin{equation}
\label{equation_wbarbart}
\begin{aligned}
\bar{\bar{\mathbf{w}}}^{(t)} &=\bar{\bar{\mathbf{w}}}^{(0)}-\frac{1}{M} \sum_{m=1}^{M} \sum_{j=0}^{t_{m}-1} \mathbbm{1}\left\{j+1 \in \mathcal{I}_{m}\right\} \mathbf{g}_{m}^{(t)} \\
&=\bar{\bar{\mathbf{w}}}^{(t_{0}^{\prime})}-\frac{1}{M} \sum_{m=1}^{M} \sum_{j=t_{0}^{\prime}}^{t_{m}-1} \mathbbm{1}\left\{j+1 \in \mathcal{I}_{m}\right\} \mathbf{g}^{(j)}_{m}
\end{aligned} 
\end{equation}

By applying Jensen's inequality, using $\left\|\mathbbm{1}\left\{j+1 \in \mathcal{I}_{m}\right\} \mathbf{g}^{(j)}_{m}\right\|^{2} \leq\left\|\mathbf{g}^{(j)}_{m}\right\|^{2}$, and taking expectation, we can upper-bound (\ref{equation_wbarbart}) as

\begin{equation}
\mathbb{E}\left\|\bar{\bar{\mathbf{w}}}^{(t_{0}^{\prime})}-\bar{\bar{\mathbf{w}}}^{(t)}\right\|^{2} \leq \frac{\left(t_{m}-t_{0}^{\prime}\right)}{M} \sum_{m=1}^{M} \sum_{j=t_{0}^{\prime}}^{t_{m}} \mathbb{E}\left\|\mathbf{g}^{(j)}_{m}\right\|^{2}
\end{equation}

Using the bound on $\mathbb{E}\left\|\mathbf{g}^{(j)}_{m}\right\|^{2}$'s from (\ref{equation_whattrr_minus_wbarbart0prime}) gives

\begin{equation}
\mathbb{E}\left\|\bar{\bar{\mathbf{w}}}^{(t_{0}^{\prime})}-\bar{\bar{\mathbf{x}}}^{(t)}\right\|^{2} \leq 32 B\left(1+\frac{C}{\gamma^{2}}\right) (\eta^{(t)})^{2} H^{4} G^{2}
\end{equation}

To bound the last term of (\ref{equation_3_bounds}), note that

\begin{equation}
\label{equantion_wtidlemtm}
\widetilde{\mathbf{w}}^{(t_{m})}_{m}=\bar{\bar{{\mathbf{w}}}}^{(0)}-\sum_{j=0}^{t_{m}-1} (\eta^{(j)}) \nabla f_{\mathcal{D}_{m}^{(j)}}\left(\widehat{\mathbf{w}}_{m}^{(j)}\right)    
\end{equation}

From (\ref{equation_wbarbart}) and (\ref{equantion_wtidlemtm}), we can write

\begin{equation}
\label{equation_wbbt_minus_1msum}
\begin{aligned}
&\bar{\bar{\mathbf{w}}}^{(t)}-\frac{1}{M} \sum_{m=1}^{M} \widetilde{\mathbf{w}}^{(t_{m})}_{m}
\\&=\frac{1}{M} \sum_{m=1}^{M}\left[\sum_{j=0}^{t_{m}-1} \eta^{(j)} \nabla f_{\mathcal{D}^{(j)}_{m}}\left(\widehat{\mathbf{w}}^{(j)}_{m}\right)-\sum_{j=0}^{t_{m}-1} \mathbbm{1}\left\{j+1 \in \mathcal{I}_{m}\right\} \mathbf{g}^{(j)}_{m}\right] 
\end{aligned}
\end{equation}

Let $t_{m}^{(1)}$ and $t_{m}^{(2)}$ be two consecutive synchronization steps in $\mathcal{I}_{m}$. Then, by the update rule of $\widehat{\mathbf{w}}^{(t)}_{m}$, we have $\widehat{\mathbf{w}}^{(t_{m}^{(1)})}_{m}-\widehat{\mathbf{w}}^{(t_{m}^{(2)}-\frac{1}{2})}_{m}=\sum_{j=t_{m}^{(1)}}^{t_{m}^{(2)}-1} \nabla f_{\mathcal{D}^{(j)}_{m}}\left(\widehat{\mathbf{w}}^{(j)}_{m}\right)$. Since $\mathbf{w}^{(t_{m}^{(1)})}_{m}=\widehat{\mathbf{w}}^{(t_{m}^{(1)})}_{m}$ and the devices do not
modify their local $\mathbf{w}^{(t)}_{m}$'s in between the synchronization steps, we have $\mathbf{w}^{(t_{m}^{(2)}-1)}_{m}=\mathbf{w}^{(t_{m}^{(1)})}_{m}=\widehat{\mathbf{w}}^{(t_{m}^{(1)})}_{m}$. Therefore, we can write

\begin{equation}
\label{equation_wm_minus_wm}
\mathbf{w}^{(t_{m}^{(2)}-1)}_{m}-\widehat{\mathbf{w}}^{(t_{m}^{(2)}-\frac{1}{2})}_{m}=\sum_{j=t_{m}^{(1)}}^{t_{m}^{(2)}-1} \nabla f_{\mathcal{D}^{(j)}_{m}}\left(\widehat{\mathbf{w}}_{m}^{(j)}\right)
\end{equation}

Using (\ref{equation_wm_minus_wm}) for every consecutive synchronization steps, we can equivalently write (\ref{equation_wbbt_minus_1msum}) as

\begin{equation}
\begin{aligned}
&\bar{\bar{\mathbf{w}}}^{(t)}-\frac{1}{M} \sum_{m=1}^{M} \widetilde{\mathbf{w}}^{(t_{m})}_{m}\\ &=\frac{1}{M} \sum_{m=1}^{M}\left[\sum_{j: j+1 \in \mathcal{I}_{m},j\leq t_m-1}\left(\mathbf{w}^{(j)}_{m}-\widehat{\mathbf{w}}^{(j+\frac{1}{2})}_{m}-\mathbf{g}^{(j)}_{m}\right)\right] \\
&=\frac{1}{M} \sum_{m=1}^{M} \mathbf{e}^{(t_{m})}_{m} \\
&=\frac{1}{M} \sum_{m=1}^{M} \mathbf{e}^{(t)}_{m}
\end{aligned}
\end{equation}

In the last inequality, we used the fact that the devices do not update their local memory in between the synchronization steps. For the reasons given in the proof of Lemma \ref{lemma_contracting_local_sequence_deviation}, we can directly apply Lemma 4 in \cite{basu2019qsparse} to bound the local memories and obtain $\mathbb{E}\left\|\frac{1}{M} \sum_{m=1}^{M} \mathbf{e}^{(t)}_{m}\right\|^{2} \leq$ $\frac{1}{M} \sum_{m=1}^{M} \mathbb{E}\left\|\mathbf{e}^{(t)}_{m}\right\|^{2} \leq 4 C \frac{(\eta^{(t)})^{2}}{\gamma^{2}} G^{2} H^{2}$. This implies

\begin{equation}
\mathbb{E}\left\|\bar{\bar{\mathbf{w}}}^{(t)}-\frac{1}{M} \sum_{m=1}^{M} \widetilde{\mathbf{w}}^{(t_{m})}_{m}\right\|^{2} \leq 4 C \frac{(\eta^{(t)})^{2}}{\gamma^{2}} G^{2} H^{2} 
\end{equation}

Putting the bounds from $(89),(92)$, and $(97)$ in $(88)$ and using $B=(4-2 \gamma)$ give

\begin{equation}
\begin{aligned}
\mathbb{E}\left\|\widehat{\mathbf{w}}^{(t)}-\widetilde{\mathbf{w}}^{(t)}\right\|^{2} &\leq 192(4-2 \gamma)\left(1+\frac{C}{\gamma^{2}}\right) (\eta^{(t)})^{2} H^{4} G^{2} \\
&+12 C \frac{(\eta^{(t)})^{2}}{\gamma^{2}} G^{2} H^{2}
\end{aligned}
\end{equation}

This completes the proof of Lemma \ref{lemma_contracting_distance_between_virtual_and_true_sequence} .

\end{document}